\documentclass[10pt,twocolumn,letterpaper]{article}

\usepackage[pagenumbers]{cvpr} % To force page numbers, e.g. for an arXiv version

\newcommand{\va}{\bm{a}}
\newcommand{\vp}{\bm{p}}

\newcommand{\vo}{\bm{o}}

\usepackage[utf8]{inputenc} % allow utf-8 input
\usepackage[T1]{fontenc}    % use 8-bit T1 fonts
\usepackage{array}          % for advanced table column formatting
\usepackage{booktabs} % 用于更好的表格线条
\usepackage{url}            % simple URL typesetting
\usepackage{booktabs}       % professional-quality tables
\usepackage{amsfonts}       % blackboard math symbols
\usepackage{bm}            % bold math symbols
\usepackage{nicefrac}       % compact symbols for 1/2, etc.
\usepackage{microtype}      % microtypography
\usepackage{xcolor}         % colors
\usepackage{amsmath,graphicx,amssymb,mathtools,amsthm,float,mathtools,wrapfig}
\usepackage{subcaption}
\usepackage{multirow}
\usepackage{wrapfig}
\usepackage{makecell}
\usepackage{dsfont}
\usepackage{algorithm}
\usepackage{algorithmic}
\usepackage{tabularx}

\usepackage[utf8]{inputenc}
\usepackage{amssymb}

\definecolor{cvprblue}{rgb}{0.21,0.49,0.74}
\usepackage[pagebackref,breaklinks,colorlinks,allcolors=cvprblue]{hyperref}

\usepackage[utf8]{inputenc}

\usepackage{mathtools}
\title{Motus: A Unified Latent Action World Model}

\author{
Hongzhe Bi$^{1*\dagger}$, Hengkai Tan$^{1*\dagger}$, Shenghao Xie$^{2,1*}$, Zeyuan Wang$^{1*}$, Shuhe Huang$^{1*}$, Haitian Liu$^{1*}$,\\
Ruowen Zhao$^{1}$, Yao Feng$^{1}$, Chendong Xiang$^{1}$, Yinze Rong$^{1}$, Hongyan Zhao$^{1}$,
Hanyu Liu$^{2}$,\\ Zhizhong Su$^{3}$, Lei Ma$^{2}$, Hang Su$^{1}$, Jun Zhu$^{1}$
\\[0.4em]
$^{1}$Dept. of Comp. Sci. and Tech., Institute for AI, BNRist Center, THBI Lab,\\
Tsinghua-Bosch Joint ML Center, Tsinghua University\\
$^{2}$Peking University\quad
$^{3}$Horizon Robotics\\[0.2em]
$^{*}$Joint first authors $^{\dagger}$Joint project lead\\
[0.2em]
\small\texttt{\{bhz24, thj23\}@mails.tsinghua.edu.cn,}
\small\texttt{dcszj@tsinghua.edu.cn}\\
[0.2em]
\textbf{Project Page:} \url{https://motus-robotics.github.io/motus} \\
}

\usepackage{float}

\begin{document}
\maketitle
%\blfootnote{$^{*}$Equal contribution}%
%\blfootnote{$^{\dagger}$Project co-lead}

\begin{abstract}
     While a general embodied agent must function as a unified system, current methods are built on isolated models for understanding, world modeling, and control. This fragmentation prevents unifying multimodal generative capabilities and hinders learning from large-scale, heterogeneous data. In this paper, we propose \textbf{Motus}, a unified latent action world model that leverages existing general pretrained models and rich, sharable motion information. Motus introduces a Mixture-of-Transformer (MoT) architecture to integrate three experts (\ie, understanding, video generation, and action) and adopts a UniDiffuser-style scheduler to enable flexible switching between different modeling modes (\ie, world models, vision-language-action models, inverse dynamics models, video generation models, and video-action joint prediction models). Motus further leverages the optical flow to learn latent actions and adopts a recipe with three-phase training pipeline and six-layer data pyramid, thereby extracting pixel-level ``delta action'' and enabling large-scale action pretraining. Experiments show that Motus achieves superior performance against state-of-the-art methods in both simulation (a \textbf{+15\%} improvement over X-VLA and a \textbf{+45\%} improvement over $\pi_{0.5}$) and real-world scenarios(improved by \textbf{+11}\textasciitilde{}\textbf{48\%}), demonstrating unified modeling of all functionalities and priors significantly benefits downstream robotic tasks.
\end{abstract}
\section{Introduction}
% para 1: EAI，理解-预测-行动闭环，统一模型至关重要
% para 2: UWM提出了一种原型，联合建模fdm, pm, idm和vgm四种范式，但没有使用预训练权重和在大规模机器人操作轨迹上训练，严重缺乏互联网的通用多模态先验和领域特定的先验。尽管video policy在svd上微调，但由于action和video采样的时间步保持一样，无法支持policy model，且没能使用vision-language understanding的先验。F1和UP-VLA虽然显式地生成未来图像，没有action作为条件，虽然能融合VLA和idm，但不支持world model，且缺乏视频级别长时序建模的能力。能否引入大规模先验来构建统一具身基础模型，both通用和领域特定的先验，实现可扩展和泛化的policy learning？
% para 3: 在这篇文章中，我们提出motus，一个unfied latent action world model。
% para 4: 挑战1：引入通用先验，vlm和vgm割裂发展，不存在一个同时强大和统一的模型。MoT+UniDiffuser
% para 5: 挑战2：引入专用先验，特定本体数据少，跨本体如何对齐。各类机器人轨迹、跨本体异构数据、motion、latent action
% contributions

\begin{figure*}[h]
  \centering
  % \fbox{\rule{0pt}{2in} \rule{0.9\linewidth}{0pt}}
   \includegraphics[width=0.7\linewidth]{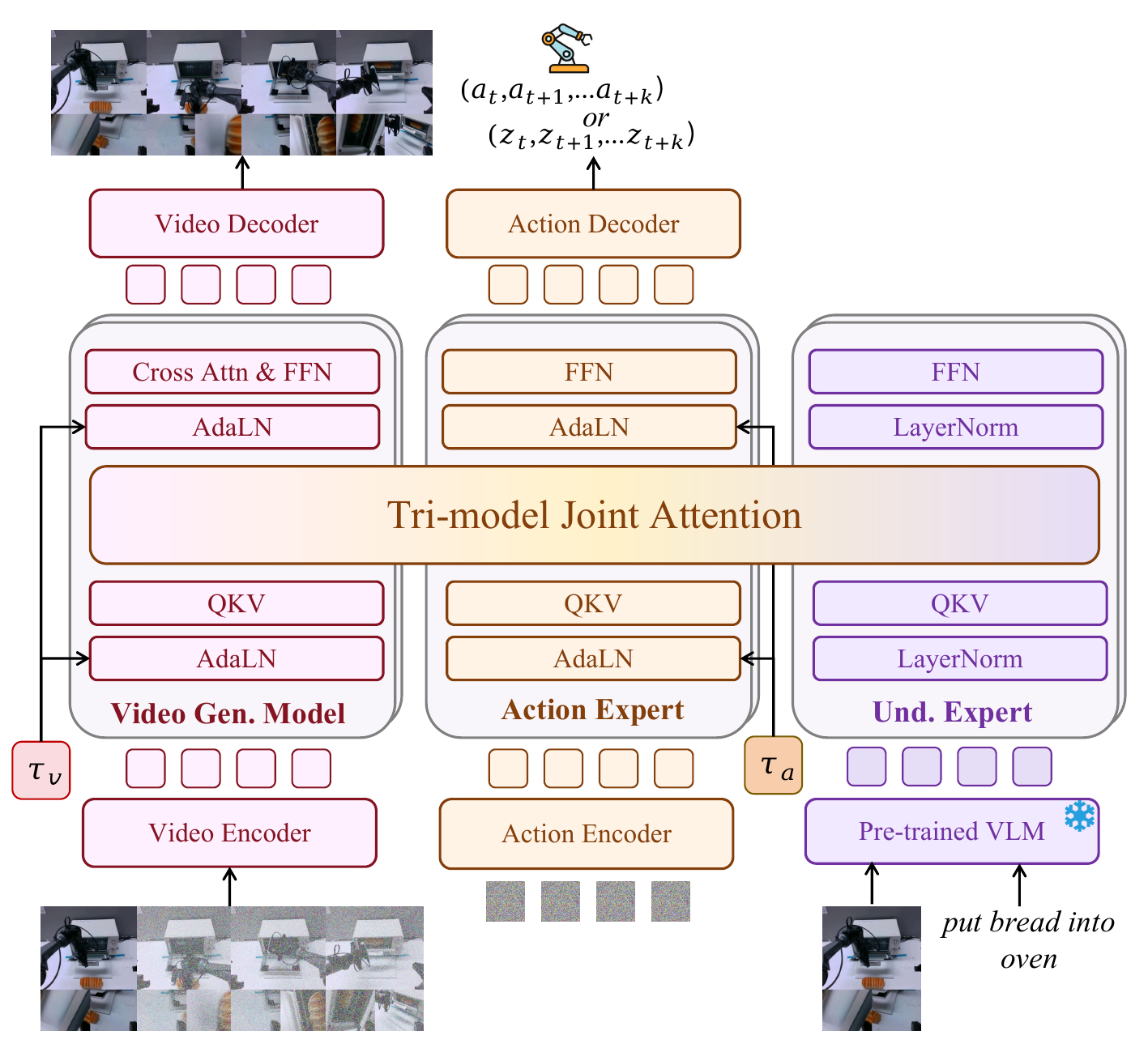}

   \caption{Motus Architecture. Here, $a_t \dots a_{t+k}$ are actions, $z_t \dots z_{t+k}$ are latent actions, and $\tau_v$ and $\tau_a$ are the rectified flow timesteps for the video generation model and the action expert, respectively.
    }
   \label{fig:motus}
\vspace{-0.3cm}
\end{figure*}
% \junz{include necessary references}\junz{make the four paradigms to be unified more explicit here}

A unified model is essential for embodied agents to integrate a spectrum of cognitive functions---from understanding scenes and instructions, imagining possible futures, to predicting consequences and generating actions---into a unified whole.
However, existing methods model these capabilities in isolation: some rely on vision-language-action models (VLAs)~\cite{black2025pi0.5, zheng2025x_vla, rt2, rtx_openx, kimopenvla, bu2025univla, liu2024rdt, bi2025hrdt} to learn static policies from vision and language; others use world models or generative approaches built on predicted futures~\cite{unisim, zhou2024robodreamer, du2023unipi, feng2025vidar, gen2act, vpp, tan2025anyposautomatedtaskagnosticactions, seer, susie, uva, video2policy}; and $\mathcal{F}_1$ \cite{lv2025f1} combines VLAs and inverse dynamics models (IDMs) by explicitly imagining future visual observations, but it excludes world models or video generation models (VGMs), resulting in incomplete unification. These approaches fragment what should be a unified system into 5 separate modeling tasks:
\begin{itemize}
    \item VLA: $p(\va_{t+1:t+k} \mid \vo_t, \ell).$
    \item WM: $p( \vo_{t+1:t+k} \mid \vo_t, \va_{t+1:t+k}).$
    \item IDM: $p( \va_{t+1:t+k} \mid \vo_{t:t+k} ).$
    \item VGM: $p( \vo_{t+1:t+k} \mid \vo_t, \ell).$
    \item Video-Action Joint Prediction Model: \\ 
    \hspace*{2.5em} $p(\vo_{t+1:t+k}, \va_{t+1:t+k} \mid \vo_t, \ell).$
\end{itemize}

Two fundamental challenges (detailed in Sec.~\ref{sec:preliminary}) hinder the integration of these capabilities. First, \textit{unifying such multimodal generative capabilities} within one framework is nontrivial. While unified world models (UWMs)~\cite{zhu2025uwm} offer a theoretical prototype, they are typically trained from scratch or with limited priors, lacking either robust vision-language understanding from vision-language models (VLMs) or rich physical interaction knowledge from VGMs. 
Second, embodied intelligence demands the ability to \textit{learn from large-scale heterogeneous data}---including internet videos, egocentric human demonstrations, and multi-robot trajectories---but action spaces vary widely across embodiments, and most video data lack action labels, making it difficult to pretrain action experts with general motion and interaction priors.

To address these challenges, we propose \textbf{Motus}, a unified latent action world model that integrates pretrained experts within a Mixture-of-Transformers (MoT) architecture. Our approach unifies the 5 key distributions by connecting a video generator (generative expert), an action expert, and a vision-language understanding expert via shared multi-head self-attention layers---a design we term \textbf{Tri-model Joint Attention}---which preserves specialized functionalities while enabling cross-modal knowledge fusion. 
To further coordinate multimodal generation, Motus incorporates a UniDiffuser-like scheduler, allocating distinct timesteps and noise scales to each modality (\eg, videos and actions). 
This enables a unified manner for simultaneous modeling marginal, conditional, and joint distributions, as well as adaptive switching among different inference modes (\eg, VLA, WM, IDM, VGM, Video-Action Joint Prediction Model). 

Additionally, to leverage heterogeneous data at scale, we introduce latent actions, which encode motion patterns from optical flow as a pixel-level “delta action”. This representation bridges visual dynamics with control signals, enabling the action expert to be pretrained on diverse unlabeled videos and robot trajectories.
Specifically, a pretrained deep compression autoencoder (DC-AE) with additional lightweight downsampling modules is used to reconstruct optical flow, whereas its encoded low-dimensional latents are supervised with a few action labels, both task-related and task-agnostic, thus steering the focus towards patterns associated with robotic activities.

Subsequently, Motus undergoes a three-phase pretraining–finetuning pipeline (\ie, video pretraining, latent action pretraining, and embodiment-specific action finetuning) on a six-layer data pyramid spanning web-scale, egocentric human, simulation, task-agnostic, multi-robotic, and target-robotic data. This recipe aligns behaviors across different embodiments within the motion space described by optical flows and shares such interaction knowledge with target embodiments to enhance the generalization in downstream tasks, thereby providing the action expert with pretraining like other experts.

Overall, our contributions can be summarized as follows: 
\begin{itemize}
    \item A unified embodied foundation model that integrates five mainstream paradigms (\ie, WMs, IDMs, VLAs, VGMs, and Video-Action Joint Prediction Models) without compromising general multimodal priors.
    \item A scalable robotic recipe with a three-phase training pipeline and six-layer data pyramid that leverages optical flow-based latent action to learn cross-embodiment transferable motion knowledge. 
    \item Extensive experiments show that Motus significantly outperforms state-of-the-art approaches in both simulation (a \textbf{+15\%} improvement over X-VLA~\cite{zheng2025x_vla} and a \textbf{+45\%} improvement over $\pi_{0.5}$~\cite{black2025pi0.5}) and real-world scenarios (improved by \textbf{+11\textasciitilde{}48\%}), demonstrating that large-scale general and domain-specific priors can be effectively fused to enhance the generalization of policy learning.
\end{itemize}

\section{Related Works}
\subsection{Unified Multimodal Models}
Unified multimodal models jointly model various modalities and tasks within a single generative framework~\cite{wang2024emu3, team2024chameleon, yang2025mmada, li2025dualdiffusion, xie2025showo2, wu2025janus}, showing broad applications across several domains~\cite{ning2025unimedvl, zhou2025hermes, ye2025shapellmomni}. In particular, Bagel~\cite{deng2025emerging_bagel} achieves unification via MoT~\cite{liang2025mot}, sharing the multi-head self-attention layers between understanding experts and generation experts. In contrast, existing embodied foundation models are developed independently, spawning multiple disparate paradigms: some leverage the text-image understanding capabilities of VLMs to learn action prediction~\cite{kim2024openvla, black2025pi0.5, bjorck2025gr00tn1}, while others utilize VGMs to generate video sequences and infer actions from consecutive frames~\cite{feng2025vidar, du2023unipi, zhou2024robodreamer}. Recently, $\mathcal{F}_1$~\cite{lv2025f1} extends VLAs to explicitly imagine future visual states and output actions by IDMs, thereby merging both models. Furthermore, UWM~\cite{zhu2025uwm} unifies WMs, VLAs, IDMs, VGMs, and Video-Action Joint Prediction Models within a single diffusion backbone, making an initial exploration of complete robotic models. Unlike UWM, our method goes beyond unified modeling by further incorporating internet-scale general multimodal priors and specialized priors from massive robotic trajectories.

\subsection{Latent Action Models}
Latent actions mitigate the scarcity of action labels by capturing visual dynamics, and are typically derived by coupling IDMs with forward dynamics models (FDMs) to reconstruct the next frame conditioned on the previous one~\cite{rybkin2018learning, edwards2019imitating, bruce2024genie, bu2025go1}. Initially, RGB images are used for supervision, but this introduces task-irrelevant appearance information~\cite{zhang2025latent}. To remove such interference, a common approach is restricting autoencoder's capacity to encode low-dimensional latents~\cite{ye2025lapa, schmidt2024lapo, chen2024moto}, thereby reducing the inclusion of redundancy. AdaWorld~\cite{gao2025adaworld} attempts to decouple the representations, such as $\beta$-VAE~\cite{higgins2017betavae}, in order to retain only the useful factors. Other approaches explore alternative reconstruction objectives, \eg, DINOv2 features~\cite{bu2025univla, chen2024moto, yang2025como}, object keypoints~\cite{yang2025tramoe, collins2025amplify, yuan2025motiontrans}, and language instructions~\cite{clark2025rad}, which carries rich semantic and spatial features. Moreover, LAOM~\cite{nikulin2025latent} employs a few action labels to encourage the model to focus on robotic activities. Building on these advances and inspired by optical flow as a universal motion expression~\cite{chefer2025videojam, wang2025lps, zhong2025flowvla}, we use it to align cross-embodiment behaviors and learn latent actions to facilitate large-scale pretraining.
\section{Problem Formulation and Challenges}
% \xsh{consider giving the data distributions of vla, wm, idm, vgm?}
\label{sec:preliminary}
\paragraph{Embodied Policies}
We consider the task of language-conditioned robotic manipulation. For each embodiment, the task defines an action $\va \in \mathcal{A}$, an observation $\vo \in \mathcal{O}$ (visual input), a language instruction $\ell \in \mathcal{L}$, and the proprioception of the robot $\vp$, where $\mathcal{A}$, $\mathcal{O}$ and $\mathcal{L}$ denote the action space, the observation space, and the language instruction space respectively.
The task typically provides an expert dataset $D_{\text{expert}} = \{\{\ell, \vp_1, \vo_1, \va_1, \dots, \vp_N, \vo_N, \va_N \}\}$, which contains robot proprioception, visual observations, and actions collected by an expert over $N$ timesteps, along with corresponding language annotations for each trajectory. 
We train a policy parameterized by $\theta$ on $D_{\text{expert}}$. At each timestep $t$, the policy predicts the next $k$ actions (action chunking~\cite{zhao2023learning_ACT}) based on the current observation and proprioception, modeling the distribution $p_\theta(\va_{t+1:t+k} \mid \vo_t, \vp_t, \ell)$ or $p_\theta(\va_{t+1:t+k} \mid \vo_t, \ell)$.
The policy $p_\theta$ is trained to maximize the likelihood objective: 
\begin{equation}
    \max_\theta \ \mathbb{E}_{(\vo_t, \vp_t, \va_{t+1:t+k}, \ell) \sim D_{\text{expert}}} \log p_\theta(\va_{t+1:t+k} \mid \vo_t, \vp_t, \ell).
\end{equation}

% \thk{5 probability distribution}   
Furthermore, based on the symbolic definitions above, we can derive the probability distributions for the 5 modeling types of embodied intelligence, which can be integrated into a single model for training:
\begin{itemize}
    \item VLA: $p(\va_{t+1:t+k} \mid \vo_t, \ell).$
    \item WM: $p( \vo_{t+1:t+k} \mid \vo_t, \va_{t+1:t+k}).$
    \item IDM: $p( \va_{t+1:t+k} \mid \vo_{t:t+k} ).$
    \item VGM: $p( \vo_{t+1:t+k} \mid \vo_t, \ell).$
    \item Video-Action Joint Prediction Model: \\ 
    \hspace*{2.5em} $p(\vo_{t+1:t+k}, \va_{t+1:t+k} \mid \vo_t, \ell).$
\end{itemize}
\paragraph{Challenge 1: Unifying Multimodal Generative Capabilities.}
A capable embodied agent must integrate a spectrum of cognitive functions---from understanding scenes and instructions, imagining possible futures, to predicting consequences and generating actions---to possess a human-like capacity, as a unified whole.
Current models, however, are fragmented and fail to capture the full set of necessary capabilities within one system.
This presents a challenge: how to unify the modeling of five key distributions---VLA, World Model, IDM, Video Generation Model, and Video-Action Joint Prediction Model---within a single framework. 
While prior work, such as UWMs~\cite{zhu2025uwm}, has made some progress, 
a critical limitation persists: 
these approaches are either trained from scratch, built upon smaller base models, or---even when incorporating some priors---invariably lack the full spectrum of knowledge, missing \textit{either} visual understanding priors from VLMs \textit{or} physical interaction priors from VGMs.
Consequently, they lack the comprehensive world knowledge required for robust and generalizable embodied intelligence.
Therefore, the nontrivial challenge of jointly modeling various distributions of vision, language, and action within a unified framework remains unaddressed, which is precisely the gap our work fills.

\paragraph{Challenge 2: Utilization of Heterogeneous Data.}
A central challenge in embodied intelligence is how to make effective use of large scale heterogeneous data. Action spaces vary widely between embodiments in dimension, range, and semantics, and robots differ in morphology, actuation, and sensing. As a result, control signals are not directly reusable and policies struggle to learn universal priors that transfer across embodiments.
Existing approaches, including~\cite{liu2024rdt, black2025pi0.5, zheng2025x_vla, WangCZH24}, try to address this by using a general backbone with embodiment-specific information injection, or constructing high-dimensional action vectors that forcibly unify different embodiments However, they still depend primarily on labeled robotic trajectories and cannot integrate these datasets with large-scale internet videos or egocentric human videos, which lack action annotations but contain abundant motion and physical interaction cues. This limitation prevents large-scale pretraining of the action expert and reduces the ability to learn general motion priors.

\section{Methodology}

\subsection{Motus}
% \thk{TODO merge these problems and challenges to sec.3}
% Current VGMs offer priors for motion and spatiotemporal information in videos, while VLMs provide priors for visual and textual understanding as well as planning. However, both lack priors for vision–language–action alignment relevant to robotic tasks, limiting their ability to guide physical robot actions.
% For robots to reach practical levels of performance, models must be capable of understanding scenes (via internet-scale image data), interpreting instructions (via language data), predicting action consequences (via world models or FDMs), imagining future task completion (via video data), and outputting precise actions. Together, these form a complete, human-like autonomous decision-making system.
% However, this problem is challenging. Naive unified training requires large-scale datasets of the form $\{\ell, \vo_1, \va_1, \dots, \vo_N, \va_N \}$, which are costly and time-consuming to collect and difficult to scale. Moreover, heterogeneous data—such as web images, videos, action-annotated robot trajectories, task-agnostic actions, and language instructions—exhibit significant modal discrepancies, making unified modeling difficult.

% To address this, we propose Motus, a unified latent action world model. 
% % To address this, we propose Motus, a unified latent action world model. Our approach leverages the pretrained capabilities of VLMs and VGMs and unifies heterogeneous data and various distributions.
% a UniDiffuser-style scheduler is also employed to assign distinct timesteps and noise scales to each modality
\paragraph{Model Architecture.} 
To address the challenges of unifying multimodal generative capabilities outlined in Sec .~\ref{sec:preliminary}, we propose Motus, a unified latent action world model. 
First, Motus is designed as a general generative model that jointly learns on heterogeneous multimodal data, thereby integrating the diverse capabilities (\eg, modeling 5 distributions) of a general-purpose system within a single network. 
Second, to circumvent the need for impractical amounts of aligned multimodal data, Motus leverages the rich, pretrained priors of existing foundation models. 
It integrates a pretrained VGM (generative expert), an understanding expert with pretrained VLM, and an action expert within a Mixture-of-Transformers (MoT) architecture (as shown in Fig.~\ref{fig:motus}), effectively fusing their complementary strengths---encompassing scenes understanding, instructions interpreting, consequences prediction, future video imagination, and action planning---without requiring full end-to-end training from scratch.
Unlike Unified World Models (UWMs)~\cite{zhu2025uwm}, which simply concatenate observation tokens and action tokens and process them through a single series of $N$ UWM blocks (containing self-attention and feed-forward network (FFN) layers), our approach leverages pretrained VLMs and VGMs by adopting a MoT structure.
In our model, each expert maintains an individual Transformer module, while the multi-head self-attention layers are concatenated, \ie, \textbf{Tri-model Joint Attention}. This not only preserves distinct function roles across experts without causing task interference but also enables effective cross-modal feature fusion, encouraging diverse pretrained knowledge to complement one another. During training, Motus jointly predicts chunks of videos and actions with rectified flow-based objectives:
\begin{equation*}
l_{\text{action}}^\theta
= \mathbb{E}_{\substack{(\vo_{t:t+k}, \va_{t+1:t+k}, \ell) \sim \mathcal{D} \\
\tau_a \sim \mathcal{U}(0, T_{\tau}) \\
\epsilon_a \sim \mathcal{N}(\bm{0}, \bm{I})}}
\big\| v_a^{\theta} - (\epsilon_a - \va_{t+1:t+k}) \big\|_2^2,
\end{equation*}

\begin{equation*}
l_{\text{obs}}^\theta
= \mathbb{E}_{\substack{(\vo_{t:t+k}, \va_{t+1:t+k}, \ell) \sim \mathcal{D} \\
\tau_{o} \sim \mathcal{U}(0, T_{\tau}) \\
\epsilon_{o} \sim \mathcal{N}(\bm{0}, \bm{I})}}
\big\| v_{o}^{\theta} - (\epsilon_{o} - \vo_{t+1:t+k}) \big\|_2^2, 
\end{equation*}

\begin{equation*}
l^\theta
= l_{\text{action}}^\theta
 + l_{\text{obs}}^\theta.
\end{equation*}
where $\vo_t$ is the condition frame, $\vo_{t+1:t+k}, \va_{t+1:t+k}$ are subsequent observations and actions, $\tau_a$ and $\tau_o$ are the assigned timesteps, $\epsilon_{a}$, $\epsilon_{o}$ are the sampled Gaussian noises, , $v_{a}^{\theta}$, $v_{o}^{\theta}$ are velocity field predicted by our unified model, and $l_{action}^{\theta}$, $l_{obs}^{\theta}$ are loss of observations and actions. By allocating different timesteps and noise scales to videos and actions, respectively, Motus establishes a UniDiffuser-like scheduler to capture heterogeneous data distributions and adaptively switch between various embodied foundation models during inference (\eg, VLA, World Model, IDM, VGM, Joint Prediction).
% the understanding expert, VGM, and action expert each maintain their own set of transformer blocks. Information exchange between these experts is achieved through joint self-attention at every block, enabling effective interaction across modalities without losing their individual pre-trained capabilities.
% Our approach leverages the strengths of existing pretrained VLMs and VGMs while incorporating diverse heterogeneous data. 
% This structure holistically models all stages of robot decision-making. 
The resulting model understands scenes, follows instructions, predicts outcomes, imagines futures, and outputs actions---all within a unified multimodal architecture. %\thk{TODO}

\begin{figure}[H]
  \centering
  % \fbox{\rule{0pt}{2in} \rule{0.9\linewidth}{0pt}}
   \includegraphics[width=\linewidth]{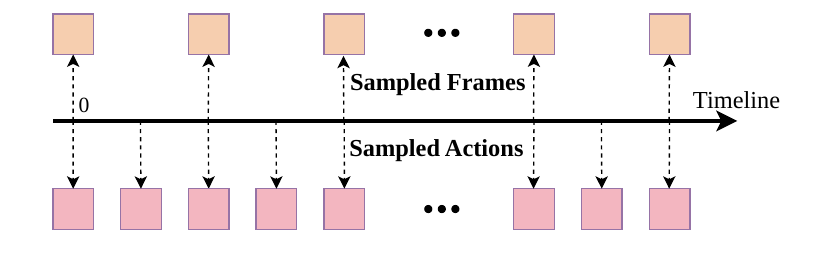}

   \caption{Action-Dense Video-Sparse Prediction. The sampling rates for video frames and actions differ.}
   \label{fig:video_action_rate}
\end{figure}
% Specifically, 
\paragraph{Action-Dense Video-Sparse Prediction.} Since our model builds upon the widely cited action-chunking technique, Motus needs to predict a chunk of future video and action sequences $\vo_{t+1:t+k}, \va_{t+1:t+k}$. This leads to several issues: (1) low training and inference efficiency, (2) redundant video frame predictions, and (3) an imbalance in the Tri-modal Joint Attention mechanism---where the number of video tokens significantly exceeds that of action tokens. This imbalance causes the model to overfit to video prediction, thereby weakening its action prediction capability. To address these problems, we propose an Action-Dense Video-Sparse Prediction strategy, as shown in Fig.~\ref{fig:video_action_rate}. During both training and inference, we downsample the video frames so that the number of video tokens and action tokens remains balanced---for example, by setting the video frame rate to one-sixth of the action frame rate.

% Specifically, Motus regulates noise levels in both the video generation model and the action expert—both implemented as diffusion models—via different timesteps $t$. 
% \thk{... unidiffuser}
\paragraph{Experts Details.} For the generative expert, we employ Wan 2.2 5B~\cite{wan2025} as the video foundation model for its accessibility and ease of use. We extend its self-attention context to create a cross-modal Tri-model Joint Attention mechanism. For the action expert, we construct a Transformer block of the same depth as Wan. Each block comprises AdaLN for injecting rectified flow timesteps, a Feed-Forward Network (FFN), and the Tri-model Joint Attention for cross-expert interaction.
We select Qwen3-VL-2B~\cite{Qwen-VL, Qwen2-VL, Qwen2.5-VL} for our understanding expert due to its inherent capabilities in 3D grounding, spatial understanding, and precise object localization, which are crucial for robotic manipulation. The input to this expert is taken from the last-layer corresponding tokens of the VLM. The understanding expert itself consists of several Transformer blocks, each containing Layer Normalization, an FFN, and the Tri-model Joint Attention.

\label{sec:model_arch}

\subsection{Latent Actions}
We further address Challenge 2 to  leverage large-scale heterogeneous data by learning generalizable action patterns directly from visual dynamics. Specifically, we introduce \textbf{latent actions} that encode the motion learned directly from pixels. These latent actions allow the model to absorb motion knowledge from various sources such as internet videos, egocentric human demonstrations, and multi-robot trajectories, thereby strengthening the pretraining of action expert even on data without explicit action labels.

\begin{figure}[H]
  \centering
  % \fbox{\rule{0pt}{2in} \rule{0.9\linewidth}{0pt}}
   \includegraphics[width=0.5\linewidth]{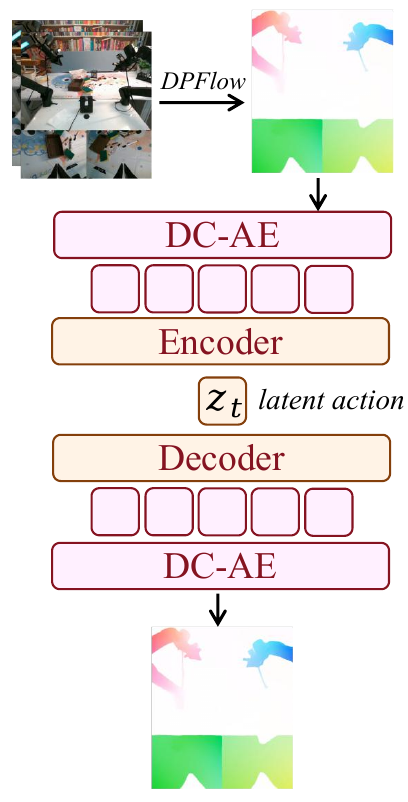}

   \caption{The Latent Action VAE.}
   \label{fig:latent_action}
\end{figure}

\paragraph{Optical Flow Based Representation.} We adopt optical flow as a natural representation of motion, which captures pixel-level displacements between consecutive frames. Specifically, optical flows are computed by DPFlow~\cite{DPFlow} and then converted into RGB images. To compress this high-dimensional representation into a control-level space, we employ a deep convolutional variational autoencoder (DC-AE~\cite{dcae}) that reconstructs the flow while encoding it into four 512-dimensional tokens. A lightweight encoder then projects these concatenated $4\times512$ features into a 14-dimensional vector, roughly matching the scale of typical robot action spaces. The overall architecture is shown in Figure~\ref{fig:latent_action}. This dimensional correspondence ensures that the latent representation can align naturally with real robotic controls and act as a bridge between perception and action.

\paragraph{Training and Distribution Alignment.} To help align the latent space to realistic action space, we incorporate task-agnostic data following AnyPos \citep{tan2025anyposautomatedtaskagnosticactions}. Specifically, task-agnostic data uses Curobo to collect image-action pairs by randomly sampling the target robot’s action space in a task-agnostic manner. This data provides additional real action supervision, helping the VAE learn an embedding that reflects feasible motor behaviors and anchors the latent actions to the true control distribution.

During training, we mix 90\% unlabeled data for self-supervised reconstruction with 10\% labeled trajectories for weak action supervision, where the labeled portion includes both task-agnostic data and standard robot demonstrations. Dimensional correspondence and weak action supervision jointly drive the latent-action distribution to align with the real action distribution, allowing motion priors learned from videos to naturally map to executable controls.

The total loss combines reconstruction, alignment, and KL regularization:

\begin{equation}
    \mathcal L=\mathcal L_{\text{recon}}+\lambda_{a}||a_{\text{real}}-a_{\text{pred}}||^2+\beta\mathcal L_{\text{KL}},
\end{equation}
where $L_{\text{recon}}$ minimizes flow-reconstruction error, the second term aligns latent and real actions, $L_{\text{KL}}$ regularizes the latent space; $\lambda_{a}$ and $\beta$ are hyperparameters.

\subsection{Model Training and Data}
\label{sec:training_data}
% Currents VGM, VLM, and non-pretrained action experts lack priors about robotic manipulation (\eg spatial...). 

% This problem leads to failures...

% However, this problem is challenging because naive unified training needs large-scale data of $\{\ell, \vo_1, \va_1, \dots, \vo_N, \va_N \}$, which is ...

% To solve this problem, our training includes 4 stages to gradually introduce manipulation-related priors into each module.

% \thk{ 
% thinking: high-level introduction of:

% Motus training (use table to clarify it ): 
% - first warmup video model in robotic dataset. only video model is trainable in this stage
% - then unified training video, latent action and language on Motus.
% - finally finetune Motus on target robot trajectory

% Data...

% }

\paragraph{Motus Training.}
Motus is trained in three structured stages (Tab.~\ref{tab:3stages}) to progressively integrate physical interaction priors from diverse datasets into a policy transferable to a target robot. Each stage addresses a key challenge:
\begin{itemize}
    \item \textbf{Stage 1: Learning Visual Dynamics.} To anchor the model in realistic physical interactions, we first adapt the Video Generation Model (VGM) using multi-robot trajectories and human videos. This enables the VGM to generate plausible future video sequences of tasks from a language instruction and an initial image.
    \item \textbf{Stage 2: Learning Action Representations.} To bridge visual forecasts with control, we pretrain the entire Motus model (VLM frozen) on videos, language, and latent actions. This stage initializes the action expert by embedding knowledge of motion and interaction into the latent action space.
    \item \textbf{Stage 3: Specializing for the Target Robot.} We finalize the model by fine-tuning it on target-robot data, ensuring that the acquired priors are fully adapted to the specific embodiment's dynamics and kinematics.
\end{itemize}

\begin{table}[h]
\centering
\caption{\textbf{Motus Training.}}
\small
\setlength{\tabcolsep}{4pt}
\renewcommand{\arraystretch}{1.2}
\begin{tabular}{>{\raggedright\arraybackslash}p{2.2cm} >{\raggedright\arraybackslash}p{3.3cm} >{\raggedright\arraybackslash}p{2cm}} % 为每一列指定宽度
\toprule
\textbf{Stage} & \textbf{Data} & \textbf{Training}  \\
\midrule
\textbf{Pretrained Foundation Models} (Off-the-shelf) & Level 1: Web Data & VGM and VLM \\ 
\midrule
\textbf{Stage 1} (Video Generation) & Level 2: Egocentric Human Videos \newline Level 3: Synthetic Data \newline Level 5: Multi-Robot Task Trajectory Data  & Only VGM \\
\midrule
\textbf{Stage 2} (Unified Training with Latent Actions) & Level 2: Egocentric Human Videos \newline Level 3: Synthetic Data \newline Level 4: Task-agnostic Data \newline Level 5: Multi-Robot Task Trajectory Data  & Motus (all 3 experts, with \textbf{latent actions}) \\
\midrule
\textbf{Stage 3} (SFT) & Level 6: Target-Robot Task Trajectory Data & Motus (all 3 experts, with actions)  \\
\bottomrule
\end{tabular}
\label{tab:3stages}
\vspace{-0.3cm}
\end{table}

\paragraph{Data.}
To equip robots with generalizable manipulation skills, we leverage large-scale multimodal data that encapsulates rich prior knowledge---from semantic understanding and physical reasoning to spatiotemporal dynamics and decision-making. As outlined in Section~\ref{sec:preliminary}, embodied data inherently spans multiple modalities: language $\ell$, image $\vo$, and action $\va$\footnote{In joint position control, proprioception and action share the same representation space.}. By considering the presence or absence of each modality, we systematically identify all meaningful data types\footnote{Language can be annotated post-hoc to support task-oriented learning.}:

\begin{itemize}
\item \textbf{Language + Image + Action}: robot trajectories (e.g., used in VLAs), $\{\ell, \vo_1, \va_1, \dots, \vo_N, \va_N \}$.
\item \textbf{Language + Image}: video sequences $\{\ell, \vo_1, \dots, \vo_N\}$ or image-text pairs $\{(\vo, \ell)\}$.
\item \textbf{Image + Action}: task-agnostic interaction data $\{(\vo_1, \va_1, \dots, \vo_{i}, \va_{i})\}$.
\item \textbf{Language-only}: textual corpora $\{\ell\}$.
\end{itemize}

\begin{figure}[H]
  \centering
  % \fbox{\rule{0pt}{2in} \rule{0.9\linewidth}{0pt}}
   \includegraphics[width=\linewidth]{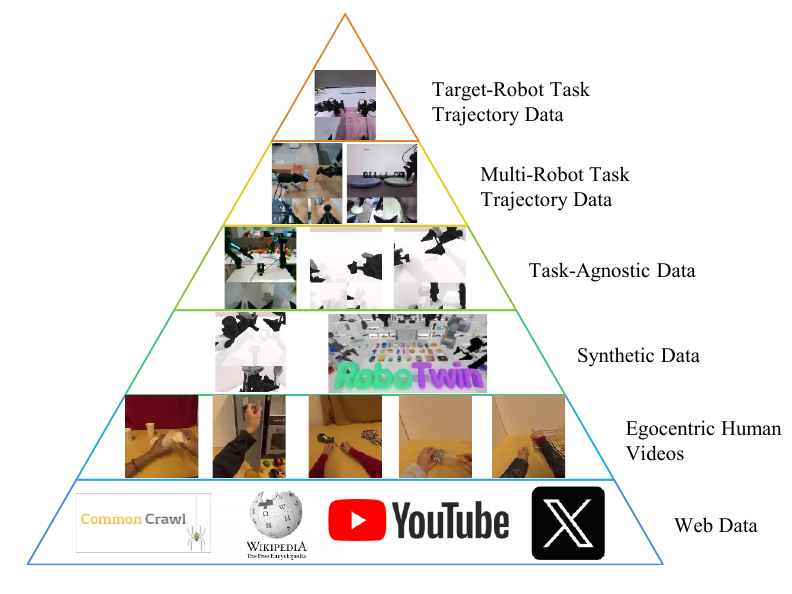}

   \caption{The Embodied Data Pyramid categorizes data into six levels, from Level 1 at the base to Level 6 at the top. Data quantity decreases from bottom to top, while data quality increases. The order of Levels 3 and 4 may sometimes vary.}
   \label{fig:data_pyramid}
\end{figure}

We exclude data lacking visual modality (e.g., language + action) as it is unsuitable for visuomotor policy learning. The remaining types form the complete spectrum of useful sources for embodied policy acquisition.
To structure this diversity, we introduce the \textit{embodied data pyramid} (Fig.~\ref{fig:data_pyramid}), which organizes data types hierarchically by richness and policy relevance.

Our framework effectively integrates and aligns all six data levels---from large-scale but indirect web sources to targeted robot demonstrations---across tailored training stages (Tab.~\ref{tab:3stages}), unifying heterogeneous datasets~\cite{contributors2025agibotworld, wu2024robomind, liu2024rdt, hoque2025egodex, chen2025robotwin} within a single, cohesive model architecture.

\vspace{-0.3cm}
\section{Experiments}
We conduct extensive experiments to assess the effectiveness of Motus in both simulated and real-world environments.
\subsection{Baselines}
We compare Motus against several state-of-the-art methods: $\pi_{0.5}$ \cite{black2025pi0.5} and X-VLA~\cite{zheng2025x_vla}. We evaluate all the models in simulation environments and further assess the performance of the baseline model $\pi_{0.5}$ in real-world tasks. We also compared both the from-scratch and Stage-1-only trained models against our own model.

\subsection{Evaluation in Simulation Environment}

\begin{table*}[ht]
  \centering
  \footnotesize
  \setlength{\tabcolsep}{5.5pt}
  \caption{\textbf{Evaluation on RoboTwin 2.0 Simulation (Clean vs Randomized, 50+ tasks).}}
  \begin{tabular}{*{1}{>{\centering\arraybackslash}m{3.2cm}} *{10}{>{\centering\arraybackslash}m{0.58cm}}}
    \toprule
    \textbf{\makecell[c]{Simulation Task}} 
      & \multicolumn{2}{c}{$\mathbf{\pi}_{\mathbf{0.5}}$}
      & \multicolumn{2}{c}{\textbf{X-VLA}}
      & \multicolumn{2}{c}{\textbf{w/o Pretrain}}
      & \multicolumn{2}{c}{\textbf{Stage1}}
      & \multicolumn{2}{c}{\textbf{Motus}} \\
    & \textbf{Clean} & \textbf{Rand.} 
    & \textbf{Clean} & \textbf{Rand.} 
    & \textbf{Clean} & \textbf{Rand.} 
    & \textbf{Clean} & \textbf{Rand.} 
    & \textbf{Clean} & \textbf{Rand.} \\
    \midrule
    \textit{Place Dual Shoes} & 12\% & 7\% & 79\% & 88\% & 78\% & 80\% & \textbf{94\%} & \textbf{94\%} & 93\% & 87\% \\
\textit{Move Stapler Pad} & 16\% & 18\% & 78\% & 73\% & 49\% & 37\% & 75\% & 68\% & \textbf{83\%} & \textbf{85\%} \\
\textit{Stack Blocks Two} & 48\% & 56\% & 92\% & 87\% & 96\% & 94\% & 99\% & \textbf{99\%} & \textbf{100\%} & 98\% \\
\textit{Scan Object} & 42\% & 38\% & 14\% & 36\% & 42\% & 50\% & 56\% & \textbf{69\%} & \textbf{67\%} & 66\% \\
\textit{Place Object Stand} & 74\% & 65\% & 86\% & 88\% & 91\% & 93\% & 93\% & 96\% & \textbf{98\%} & \textbf{97\%} \\
\textit{Place Fan} & 25\% & 36\% & 80\% & 75\% & 77\% & 85\% & 77\% & 85\% & \textbf{91\%} & \textbf{87\%} \\
\textit{Move Pillbottle Pad} & 33\% & 29\% & 73\% & 71\% & 83\% & 83\% & \textbf{96\%} & 90\% & 93\% & \textbf{96\%} \\
\textit{Pick Dual Bottles} & 10\% & 6\% & 47\% & 36\% & 58\% & 68\% & 7\% & 17\% & \textbf{96\%} & \textbf{90\%} \\
\textit{Blocks Ranking Rgb} & 43\% & 35\% & 83\% & 83\% & 92\% & 88\% & 97\% & \textbf{98\%} & \textbf{99\%} & 97\% \\
\textit{......(50 tasks)} &&&&&&&&&& \\
\textit{Turn Switch} & 5\% & 6\% & 40\% & 61\% & 69\% & 60\% & 59\% & 64\% & \textbf{84\%} & \textbf{78\%} \\
\textit{Pick Diverse Bottles} & 5\% & 3\% & 58\% & 36\% & 53\% & 62\% & 18\% & 18\% & \textbf{90\%} & \textbf{91\%} \\
\textit{Place Bread Basket} & 48\% & 56\% & 81\% & 71\% & 73\% & 83\% & 89\% & 87\% & \textbf{91\%} & \textbf{94\%} \\
\textit{Stack Blocks Three} & 15\% & 16\% & 6\% & 10\% & 71\% & 76\% & \textbf{99\%} & \textbf{95\%} & 91\% & \textbf{95\%} \\
\textit{Put Bottles Dustbin} & 12\% & 9\% & 74\% & 77\% & 36\% & 33\% & 34\% & 24\% & \textbf{81\%} & \textbf{79\%} \\
\textit{Place Can Basket} & 19\% & 25\% & 49\% & 52\% & 46\% & 62\% & 66\% & 55\% & \textbf{81\%} & \textbf{76\%} \\
\textit{Stamp Seal} & 36\% & 23\% & 76\% & 82\% & 80\% & 88\% & \textbf{93\%} & \textbf{95\%} & \textbf{93\%} & 92\% \\
\textit{Hanging Mug} & 3\% & 3\% & 23\% & 27\% & 14\% & 10\% & 37\% & 25\% & \textbf{38\%} & \textbf{38\%} \\
\textit{Handover Block} & 18\% & 19\% & 73\% & 37\% & 34\% & 15\% & 55\% & 55\% & \textbf{86\%} & \textbf{73\%} \\
\textit{Stack Bowls Three} & 33\% & 35\% & 76\% & 86\% & \textbf{90\%} & 74\% & 86\% & 83\% & 79\% & \textbf{87\%} \\
\textit{Place Object Basket} & 43\% & 36\% & 44\% & 39\% & 74\% & 75\% & 76\% & 80\% & \textbf{81\%} & \textbf{87\%} \\
\textit{Open Microwave} & 35\% & 37\% & 79\% & 71\% & 83\% & 82\% & 82\% & 84\% & \textbf{95\%} & \textbf{91\%} \\
    \midrule
    \textbf{\textit{Average (\%)}} & 42.98 & 43.84 & 72.80 & 72.84 & 72.8 & 77.00 & 82.86 & 81.86 & \textbf{88.66} & \textbf{87.02} \\
    \bottomrule
  \end{tabular}
  \label{tab:robotwin-short}
\vspace{-0.3cm}
\end{table*}

We evaluated single-task performance on 50 representative manipulation tasks from the RoboTwin 2.0 tasks in randomized scenes. To probe the general ability of our method, we carry out multi-task training: Motus and all baselines are trained on 2500 demonstrations collected in clean scenes (50 per task) plus 25000 demonstrations gathered in heavily randomized scenes (500 per task). The randomization includes random backgrounds, a cluttered table, table-height perturbations, and randomized lighting. All models are finetuned for 40k steps on the RoboTwin dataset starting from their pretrained checkpoints, and we evaluate performance by measuring the success rate of each task over 100 execution trials.

This benchmark is particularly challenging and informative because it contains a large variety of task scenes and randomized instructions, testing a model’s ability to handle various manipulation settings. Its strong background and environmental variability further evaluate the generalization under distribution shift. Moreover, all models are allowed only 40k finetuning steps on top of their pretrained checkpoints, providing a strict and fair assessment of the effectiveness of different pretraining strategies.

As shown in Tab.~\ref{tab:robotwin-short}, Motus achieves state-of-the-art performance on the RoboTwin 2.0 randomized multi-task setting, delivering over a 45\% absolute improvement compared with the $\pi_{0.5}$ model. By using a unified MoT model, Motus successfully integrates vision, language, and action generation, solving \textbf{Challenge 1}. In \textbf{Challenge 2}, the introduction of latent actions enables Motus to effectively leverage both labeled and large-scale unlabeled data, improving generalization across embodiments and capturing rich motion priors. This combination of techniques allows Motus to overcome the limitations of previous approaches and achieve superior performance.

\subsection{Real-World Experiments}
We evaluate Motus across two distinct real-world dual-arm robotic platforms, AC-One and Agilex-Aloha-2 under a comprehensive set of non-trivial tasks that span various dimensions of policy capabilities including: (1) Spatial Understanding (2) Deformable Objects Manipulation (3) Precision Fluid Control (4) Visual understanding (5) Long-Horizon Planning, such as fold towel, brew coffee using drip coffee machine and grind coffee beans with grinder.

For each task, we employed 100 trajectories for training. Consistent with the simulator, a multi-task joint training scheme was adopted: all tasks on each robotic platform were trained collectively within a single model, which was subsequently evaluated on every individual task. This approach provides a comprehensive and rigorous assessment of the model's robustness and generalization capabilities.

We choose $\pi_{0.5}$ as our baseline. Since most tasks involve long-horizon reasoning and are decomposable, we employed the partial success rate for evaluation. This metric quantifies performance by decomposing a task into subtasks, where the model earns partial scores for achieving specific subgoals and a full score only for overall success, thereby offering a more compelling demonstration of its capability. Examples are shown in Table~\ref{tab:psr-questions} and Table~\ref{tab:psr-water}.

\begin{figure}
  \centering
   \includegraphics[width=1\linewidth]{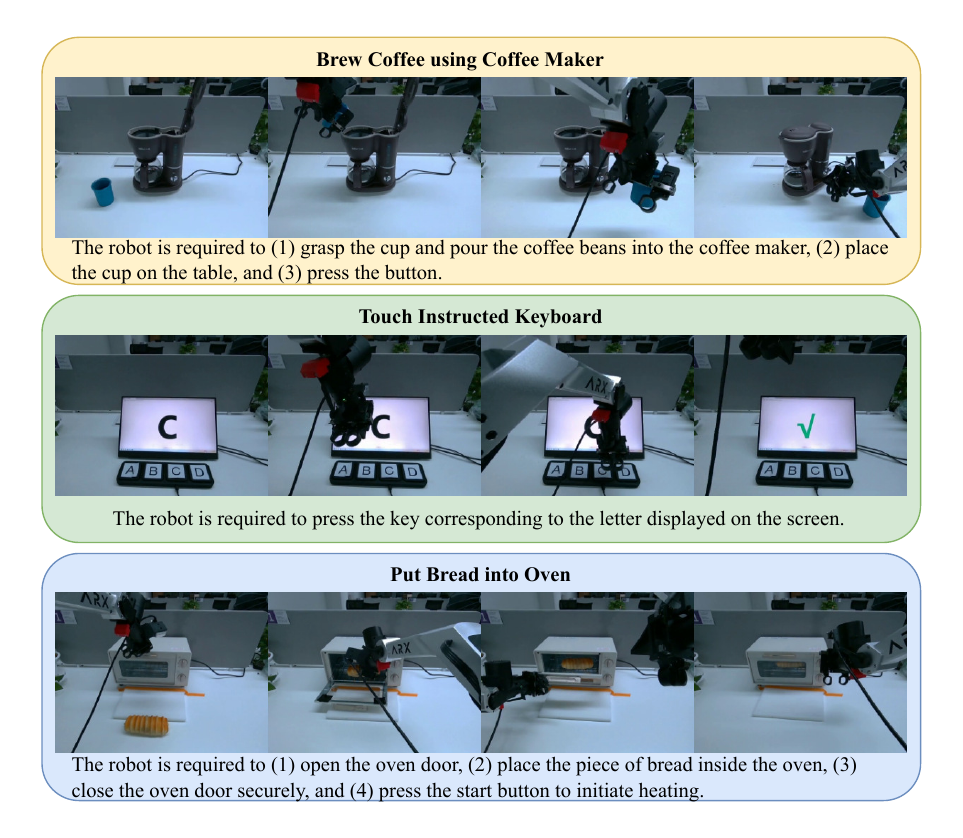}
   \caption{\textbf{Task Definitions and Visualizations.} For each task, we describe its language instruction and definitions of each sub-task.}
   \label{fig:tasks}
\end{figure}

The results are reported in Table~\ref{tab:real-world}. Our results demonstrate that Motus significantly outperforms the baseline $\pi_{0.5}$ across all tasks on both robotic arms. Visualizations are provided in Figure~\ref{fig:tasks}

\noindent

\begin{table}
\centering
  \footnotesize
\caption{Robotic Manipulation Tasks Performance Across Platforms (Partial Success Rate \%).}
\begin{tabular}{lccc}
\toprule
\textbf{Task Description} & \textbf{$\pi_{0.5}$} & \textbf{w/o Pretrain} & \textbf{Motus} \\
\midrule
\multicolumn{4}{c}{\textbf{AC-One}} \\
\midrule
Fold Towel & 4 & 1 & \textbf{14.5} \\
Brew Coffee using Coffee Maker & 0 & 0 & \textbf{62} \\
Get Water from Water Dispenser & 30 & 8 & \textbf{36} \\
Place Cube into Plate & 46 & 60 & \textbf{100} \\
Place Cube into Plate(OOD) & 28.125 & 18.75 & \textbf{75} \\
Grind Coffee Beans with Grinder & 8 & 0 & \textbf{92} \\
Pour Water from Kettle to Flowers & 5 & 5 & \textbf{65} \\
Touch Instructed Keyboard & 0 & \textbf{100} & 82.5 \\
Put Bread into Oven & 12 & 40 & \textbf{42} \\
\cline{1-1}
Average & 14.79 & 25.86 & \textbf{63.22} \\
\midrule
\multicolumn{4}{c}{\textbf{Agilex-Aloha-2}} \\
\midrule
Fold Towel & 27.5 & 0 & \textbf{39} \\
Get Water from Water Dispenser & 62 & 8 & \textbf{96} \\
Pour Water from Kettle to Flowers & 45 & 40 & \textbf{47.5} \\
Touch Instructed Keyboard & 72.5 & \textbf{85} & 80 \\
Put Bread into Oven & \textbf{36} & 0 & 34 \\
\cline{1-1}
Average & 48.60 & 26.60 & \textbf{59.30} \\
\bottomrule
\end{tabular}
\label{tab:real-world}
\vspace{-0.3cm}
\end{table}

\begin{table}
\centering
  \footnotesize
\caption{Put Bread into Oven Task on AC-One Platform with a Detailed Subtask Breakdown. The number preceding each subtask indicates the score assigned to its successful completion.}
\setlength{\tabcolsep}{4pt}
\begin{tabular}{cccc}
\toprule
\textbf{Subgoal} & $\pi_{0.5}$ & \textbf{w/o Pretrain} & \textbf{Motus} \\
\midrule
0.0: Complete Failure & 6 & 4 & 5 \\
0.2: Open the Oven & 3 & 0 & 0 \\
0.4: Grab the Bread & 0 & 2 & 1 \\
0.6: Put the Bread into the Oven & 1 & 1 & 0 \\
0.8: Close the Oven & 0 & 2 & 1 \\
1.0: Spin the Button & 0 & 1 & 3 \\
Partial Success Rate  & 12\% & 40\% & 42\% \\
\bottomrule
\end{tabular}
\label{tab:psr-questions}
\vspace{-0.3cm}
\end{table}

\begin{table}
\centering
  \footnotesize
\caption{Get Water from Water Dispenser Task on Agilex-Aloha-2 Platform with a Detailed Subtask Breakdown. The number preceding each subtask indicates the score assigned to its successful completion.}
\setlength{\tabcolsep}{4pt}
\begin{tabular}{cccc}
\toprule
\textbf{Subgoal} & $\pi_{0.5}$ & \textbf{w/o Pretrain} & \textbf{Motus} \\
\midrule
0.0: Complete Failure & 0 & 8 & 0 \\
0.4: Grab the cup & 5 & 2 & 0 \\
0.8: Fill the cup with water & 4 & 0 & 2 \\
1.0: Complete Success & 1 & 0 & 8 \\
Partial Success Rate  & 62\% & 8\% & 96\% \\
\bottomrule
\end{tabular}
\label{tab:psr-water}
\vspace{-0.3cm}
\end{table}

% \begin{table*}[h]
%   \centering
%   \footnotesize
%   \setlength{\tabcolsep}{5.5pt}
%   \caption{\textbf{Real-World Experiments.}}
%   \begin{tabular}{*{1}{>{\centering\arraybackslash}m{3.2cm}} *{16}{>{\centering\arraybackslash}m{0.38cm}}}
%     \toprule
%     \textbf{Task} & \textbf{Motus} & \textbf{Pi0.5} & \textbf{Go-1} \\
%     \midrule
%     \textit{Adjust Bottle}& XX & XX & XX \\
%     \textit{Beat Block Hammer}& XX & XX & XX  \\
%     \textit{Blocks Ranking RGB}& XX & XX & XX  \\
%     \textit{Stack Bowls Two}& XX & XX & XX  \\
%     \textit{Stamp Seal}& XX & XX & XX \\
%     \textit{Turn Switch}& XX & XX & XX  \\
%     \bottomrule
%   \end{tabular}
%   \label{tab:real-world}
% \end{table*}

\subsection{Ablation Study}
\begin{figure}
  \centering
  % \fbox{\rule{0pt}{2in} \rule{0.9\linewidth}{0pt}}
   \includegraphics[width=\linewidth]{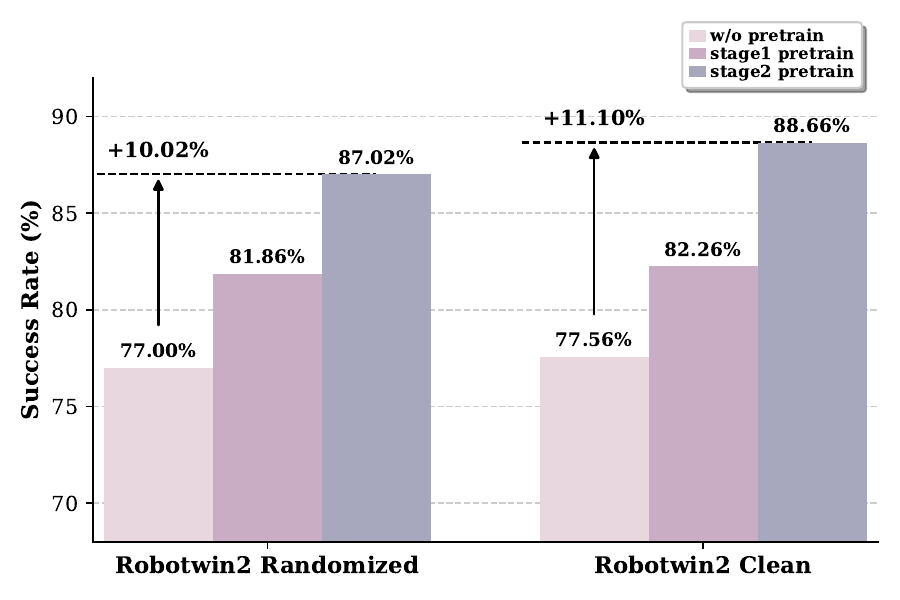}
   \caption{\textbf{Ablation in RoboTwin 2.0 Randomized Multi-task Setting}. The figure presents the total success rates (\%) of the original Motus (Stage 2 Pretrain) and its two variants: Without Pretrain and Stage 1 Pretrain.}
   \label{fig:ablation}
\vspace{-0.3cm}
\end{figure}

We performed ablation studies to demonstrate the contribution of each training stage. This involved benchmarking models without pretraining and only Stage 1 pretraining. Evaluations were carried out in the RoboTwin 2.0 simulator to measure accuracy. In real-world deployments we compare Motus against its from-scratch counterpart.
The results in simulator are summarized in Fig ~\ref{fig:ablation}, and results in real-world experiments are shown in Table~\ref{tab:real-world}. 
% \begin{table}[t]
% \centering
% \caption{\textbf{Ablation study on RoboTwin 2.0 benchmark}. The accuracy values in the second column represent the overall accuracy across the 50 tasks.}
% \small
% \setlength{\tabcolsep}{4pt}
% \begin{tabular}{cc}
% \toprule
% \textbf{Configuration} & \textbf{Acc(\%)} \\
% \midrule
% 78.3 & 26.8 \\
% 76.1 & 25.4 \\
%  75.9 & 24.2 \\
%  74.5 & 23.7 \\
% \bottomrule
% \end{tabular}
% \label{tab:ablation}
% \end{table}

\section{Conclusion and Limitations}
In this work, we present Motus, a unified latent-action world model that integrates mainstream capabilities of embodied foundation models into a single generative framework, \ie, vision-language understanding, video generation, inverse dynamics, world modeling, and video-action joint prediction. By connecting pretrained experts through MoT, coordinating multimodal modeling with a UniDiffuser-style scheduler, and introducing latent actions as a pixel-level ``delta action'' and motion representation, Motus effectively learns from large-scale heterogeneous data and inherits both general multimodal priors and rich physical interaction knowledge.
Extensive experiments across simulation and real-world environments demonstrate that Motus consistently outperforms existing state-of-the-art embodied models (improved by \textbf{+15\textasciitilde{}45\%} in simulation and \textbf{+11\textasciitilde{}48\%} in real-world scenarios), validating the importance of unifying multimodal generative capabilities and shared motion priors. We hope Motus inspires future research on unified architectures, motion-centric representation learning, and large-scale embodied pretraining.

In the future, we will continue to explore more advanced unified model architectures, pursue more universal motion priors, and learn latent actions from internet-scale general videos for embodied intelligence.

{
    \small
    \bibliographystyle{ieeenat_fullname}
    \bibliography{refs}
}

% WARNING: do not forget to delete the supplementary pages from your submission 
\clearpage
\setcounter{page}{1}
\maketitlesupplementary

\section{Training and Inference of the Unified Model}
% TODO: unidiffuser theory and inference

In this section, we analyze the training and inference procedures of the unified model, from both theoretical and experimental perspectives.

\subsection{Theorectical Analysis}
During each training iteration, given $o^0_{t:t+k}$ and $a^0_{t:t+k}$, Motus samples different timesteps $\tau_o$, $\tau_a$ and noise $\epsilon_o$, $\epsilon_a$ for them respectively, construct the interpolated trajectories $o_{t+1:t+k}^{\tau_o}$, $a_{t+1:t+k}^{\tau_a}$ based on rectified flow, and compute the loss between the predicted velocity field $v_{o}^{\theta}$, $v_{a}^{\theta}$ and its ground truth $v_{o}$, $v_{a}$ obtained by path differentiation with $t$.

% First, the unified training objective of Motus is:
% \begin{align}
% l_{\text{action}}^\theta
% &= \mathbb{E}_{\substack{(\vo_{t:t+k}, \va_{t+1:t+k}, \ell) \sim \mathcal{D} \\
% \tau_a \sim \mathcal{U}(0, T_{\tau}) \\
% \epsilon_a \sim \mathcal{N}(\bm{0}, \bm{I})}}
% \big\| v_a^{\theta} - (\epsilon_a - \va_{t+1:t+k}) \big\|_2^2, \\
% l_{\text{obs}}^\theta
% &= \mathbb{E}_{\substack{(\vo_{t:t+k}, \va_{t+1:t+k}, \ell) \sim \mathcal{D} \\
% \tau_{o} \sim \mathcal{U}(0, T_{\tau}) \\
% \epsilon_{o} \sim \mathcal{N}(\bm{0}, \bm{I})}}
% \big\| v_{o}^{\theta} - (\epsilon_{o} - \vo_{t+1:t+k}) \big\|_2^2,  \\
% l^\theta
% &= l_{\text{action}}^\theta
%  + l_{\text{obs}}^\theta.
% \end{align}

\begin{algorithm}[H]
    \caption{Training}
    \label{alg:train}
    \begin{algorithmic}[1]
        \STATE \textbf{repeat}
        \STATE\ \  $o^0_{t:t+k}, a^0_{t+1:t+k}, \ell \sim D_{expert}$
        \STATE\ \  $\tau_{o}, \tau_{a} \sim \mathrm{Uniform}(\{1,2,\dots,T_\tau\})$
        \STATE\ \  $\epsilon_o, \epsilon_a \sim \mathcal{N}(\bm{0}, \bm{I})$
        \STATE\ \  $o_{t+1:t+k}^{\tau_o} = (1-\tau_o) o^0_{t+1:t+k} + \tau_o \epsilon_o$
        \STATE\ \  $a_{t+1:t+k}^{\tau_a} = (1-\tau_a) a^0_{t+1:t+k} + \tau_a \epsilon_a$ 
        \STATE\ \  $v_{o}^{\theta}, v_{a}^{\theta} = \text{Model}_\theta(o^0_t, o_{t+1:t+k}^{\tau_o}, a_{t+1:t+k}^{\tau_a}, \tau_{o}, \tau_a, \ell)$
        \STATE\ \  $l_{\text{action}}^\theta = \| v_a^{\theta} - (\epsilon_a - a^0_{t+1:t+k}) \|_2^2$
        \STATE\ \  $l_{\text{obs}}^\theta = \| v_o^{\theta} - (\epsilon_o - o^0_{t+1:t+k}) \|_2^2$
        \STATE\ \  $l^\theta = l_{\text{action}}^\theta + l_{\text{obs}}^\theta$
        \STATE\ \  $\theta \leftarrow \theta - \eta \nabla_\theta l^\theta$
        \STATE \textbf{until} converged
    \end{algorithmic}
\end{algorithm}

During inference, Motus can switch between the following five different modes.

\paragraph{VGM.}
To enable VGM $p(o^0_{t+1:t+k} \mid o^0_t, \ell)$, given $o^0_t$ and $\ell$ as conditions, we set the starting timesteps for both the observations and actions to $T_\tau$, randomly sample $\epsilon_a, \epsilon_o \sim \mathcal{N}(\bm{0}, \bm{I})$, then apply Alg.~\ref{alg:vgm} to gradually infer $o^0_{t+1:t+k}$ from $\epsilon_o$, while keeping $a^{T_\tau}_{t+1:t+k}$ consistently noisy as $\epsilon_a$.

\begin{algorithm}[H]
    \caption{VGM}
    \label{alg:vgm}
    \begin{algorithmic}[1]
        \REQUIRE\ \  $o^0_t, \ell, \theta$
        \STATE\ \  $\epsilon_o, \epsilon_a  \sim \mathcal{N}(\bm{0}, \bm{I})$
        \STATE\ \  $o_{t+1:t+k}^{T_\tau} \leftarrow \epsilon_o$
        \STATE\ \  $a_{t+1:t+k}^{T_\tau} \leftarrow \epsilon_a$
        \STATE \textbf{for} $\tau = T_\tau \dots 1$ \textbf{do}
        \STATE\ \  $v_{o}, v_{a} = \text{Model}_\theta(o^0_t, o_{t+1:t+k}^{\tau}, a_{t+1:t+k}^{T_\tau}, \tau, T_\tau, \ell)$
        \STATE\ \  $o_{t+1:t+k}^{\tau-1} = o_{t+1:t+k}^{\tau} + v_{o}d\tau$
        \STATE \textbf{end for}
        \STATE \textbf{return} $o_{t+1:t+k}^{0}$
    \end{algorithmic}
\end{algorithm}

\paragraph{World Model.}
To enable world model $p(o^0_{t+1:t+k} \mid o^0_t, a^0_{t+1:t+k})$, given $o^0_t$ and $a^0_{t+1:t+k}$ as conditions, we set the starting timesteps for the observations and actions to $T_\tau$ and $0$ respectively, randomly sample $\epsilon_o \sim \mathcal{N}(\bm{0}, \bm{I})$, then apply Alg.~\ref{alg:wm} to gradually infer $o^0_{t+1:t+k}$ from $\epsilon_o$, while keeping $a^0_{t+1:t+k}$ always clean.

\begin{algorithm}[H]
    \caption{World Model}
    \label{alg:wm}
    \begin{algorithmic}[1]
        \REQUIRE\ \  $o^0_t, a^0_{t+1:t+k}, \ell, \theta$
        \STATE\ \  $\epsilon_o  \sim \mathcal{N}(\bm{0}, \bm{I})$
        \STATE\ \  $o_{t+1:t+k}^{T_\tau} \leftarrow \epsilon_o$
        \STATE \textbf{for} $\tau = T_\tau \dots 1$ \textbf{do}
        \STATE\ \  $v_{o}, v_{a} = \text{Model}_\theta(o^0_t, o_{t+1:t+k}^{\tau}, a^0_{t+1:t+k}, \tau, 0, \ell)$
        \STATE\ \  $o_{t+1:t+k}^{\tau-1} = o_{t+1:t+k}^{\tau} + v_{o}d\tau$
        \STATE \textbf{end for}
        \STATE \textbf{return} $o_{t+1:t+k}^{0}$
    \end{algorithmic}
\end{algorithm}

\paragraph{IDM.}
To enable IDM $p(a^0_{t+1:t+k} \mid o^0_{t:t+k})$, given $o^0_{t:t+k}$ as conditions, we set the starting timesteps for the observations and actions to $0$ and $T_\tau$ respectively, randomly sample $\epsilon_a \sim \mathcal{N}(\bm{0}, \bm{I})$, then apply Alg.~\ref{alg:idm} to gradually infer $a^0_{t+1:t+k}$ from $\epsilon_a$, while keeping $o^0_{t:t+k}$ always clean.

\begin{algorithm}[H]
    \caption{IDM}
    \label{alg:idm}
    \begin{algorithmic}[1]
        \REQUIRE\ \  $o^0_{t:t+k}, \ell, \theta$
        \STATE\ \  $\epsilon_a  \sim \mathcal{N}(\bm{0}, \bm{I})$
        \STATE\ \  $a_{t+1:t+k}^{T_\tau} \leftarrow \epsilon_a$
        \STATE \textbf{for} $\tau = T_\tau \dots 1$ \textbf{do}
        \STATE\ \  $v_{o}, v_{a} = \text{Model}_\theta(o^0_{t:t+k}, a_{t+1:t+k}^{\tau}, 0, \tau, \ell)$
        \STATE\ \  $a_{t+1:t+k}^{\tau-1} = a_{t+1:t+k}^{\tau} + v_{a}d\tau$
        \STATE \textbf{end for}
        \STATE \textbf{return} $a_{t+1:t+k}^{0}$
    \end{algorithmic}
\end{algorithm}

\paragraph{VLA.}
To enable VLA $p(a^0_{t+1:t+k} \mid o^0_t, \ell)$, given $o^0_t$ and $\ell$ as conditions, we set the starting timesteps for both the observations and actions to $T_\tau$, randomly sample $\epsilon_a, \epsilon_o \sim \mathcal{N}(\bm{0}, \bm{I})$, then apply Alg.~\ref{alg:vla} to gradually infer $a^0_{t+1:t+k}$ from $\epsilon_a$, while keeping $o^{T_\tau}_{t+1:t+k}$ consistently noisy as $\epsilon_o$.

\begin{algorithm}[h]
    \caption{VLA}
    \label{alg:vla}
    \begin{algorithmic}[1]
        \REQUIRE\ \  $o^0_t, \ell, \theta$
        \STATE\ \  $\epsilon_o, \epsilon_a  \sim \mathcal{N}(\bm{0}, \bm{I})$
        \STATE\ \  $o_{t+1:t+k}^{T_\tau} \leftarrow \epsilon_o$
        \STATE\ \  $a_{t+1:t+k}^{T_\tau} \leftarrow \epsilon_a$
        \STATE \textbf{for} $\tau = T_\tau \dots 1$ \textbf{do}
        \STATE\ \  $v_{o}, v_{a} = \text{Model}_\theta(o^0_t, o_{t+1:t+k}^{T_\tau}, a_{t+1:t+k}^{\tau}, T_\tau, \tau, \ell)$
        \STATE\ \  $a_{t+1:t+k}^{\tau-1} = a_{t+1:t+k}^{\tau} + v_{a}d\tau$
        \STATE \textbf{end for}
        \STATE \textbf{return} $a_{t+1:t+k}^{0}$
    \end{algorithmic}
\end{algorithm}

\paragraph{Video-Action Joint Prediction Model.}
To enable video-action joint prediction model $p(o^0_{t+1:t+k}, a^0_{t+1:t+k} \mid o^0_t, \ell)$, given $o_t$ and $\ell$ as conditions, we set the starting timesteps for both the observations and actions to $T_\tau$, randomly sample $\epsilon_a, \epsilon_o \sim \mathcal{N}(\bm{0}, \bm{I})$, then apply Alg.~\ref{alg:vgm} to gradually infer $a^0_{t+1:t+k}$ from $\epsilon_a$ and $o^0_{t+1:t+k}$ from $\epsilon_o$.

\begin{algorithm}[H]
    \caption{Video-Action Joint Prediction Model}
    \label{alg:joint_pred}
    \begin{algorithmic}[1]
        \REQUIRE\ \  $o^0_t, \ell, \theta$
        \STATE\ \  $\epsilon_o, \epsilon_a  \sim \mathcal{N}(\bm{0}, \bm{I})$
        \STATE\ \  $o_{t+1:t+k}^{T_\tau} \leftarrow \epsilon_o$
        \STATE\ \  $a_{t+1:t+k}^{T_\tau} \leftarrow \epsilon_a$
        \STATE \textbf{for} $\tau = T_\tau \dots 1$ \textbf{do}
        \STATE\ \  $v_{o}, v_{a} = \text{Model}_\theta(o^0_t, o_{t+1:t+k}^{\tau}, a_{t+1:t+k}^{\tau}, \tau, \tau, \ell)$
        \STATE\ \  $o_{t+1:t+k}^{\tau-1} = o_{t+1:t+k}^{\tau} + v_{o}d\tau$
        \STATE\ \  $a_{t+1:t+k}^{\tau-1} = a_{t+1:t+k}^{\tau} + v_{a}d\tau$
        \STATE \textbf{end for}
        \STATE \textbf{return} $o_{t+1:t+k}^{0}, a_{t+1:t+k}^{0}$
    \end{algorithmic}
\end{algorithm}

\subsection{Experimental Results}

\begin{figure*}
    \centering
    \includegraphics[width=0.9\linewidth]{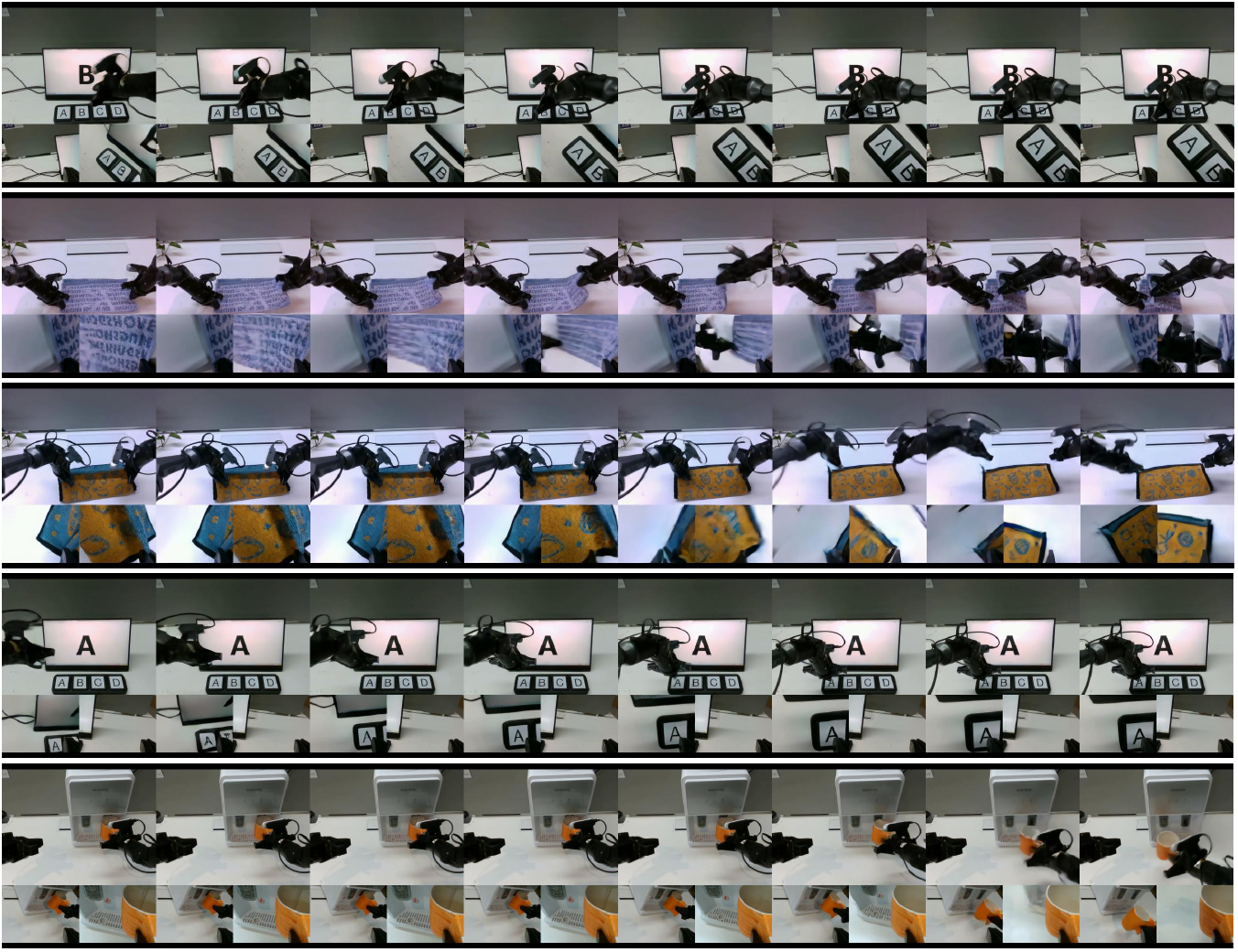}
    \caption{\textbf{Visualization of Motus's VGM mode on Agilex-Aloha-2.}}
    \label{fig:vgm_aloha}
\vspace{-0.3cm}
\end{figure*}

\begin{figure*}
    \centering
    \includegraphics[width=0.9\linewidth]{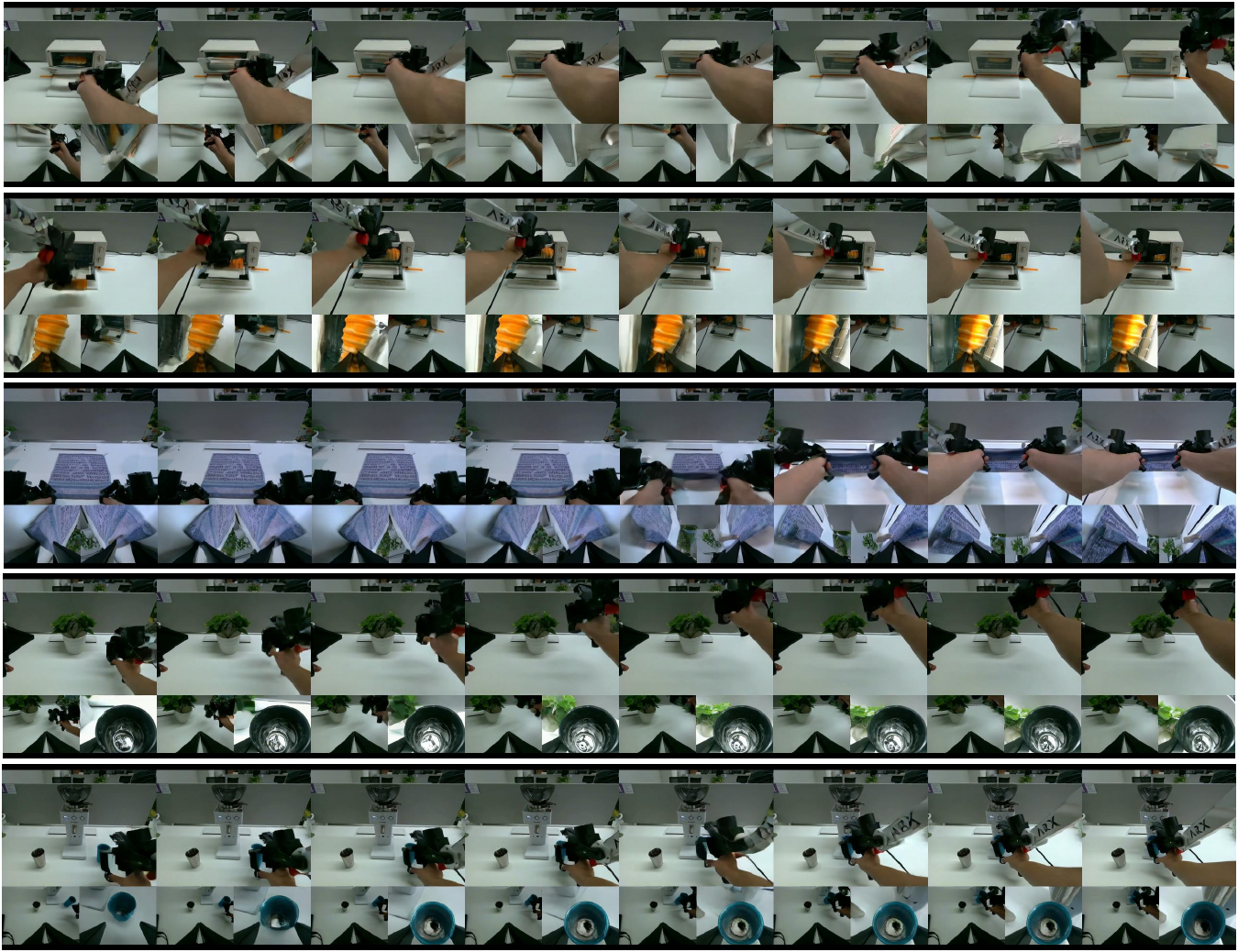}
    \caption{\textbf{Visualization of Motus's VGM mode on AC-One.}}
    \label{fig:vgm_acone}
\vspace{-0.3cm}
\end{figure*}

\paragraph{VGM.} As shown in Fig.~\ref{fig:vgm_aloha} and Fig.~\ref{fig:vgm_acone}, when Motus performs in VGM mode, it shows high-quality visualization results across both Agilex-Aloha-2 and AC-One embodiments, demonstrating the strong video generation capabilities.

% \begin{figure*}
%     \centering
%     \includegraphics[width=0.9\linewidth]{figs/vgm/vgm_acone.pdf}
%     \caption{\textbf{Visualization of Motus's VGM mode on AC-One.}}
%     \label{fig:vgm_acone}
% \vspace{-0.3cm}
% \end{figure*}

\begin{figure*}[p!]
    \centering
    \includegraphics[width=0.8\linewidth]{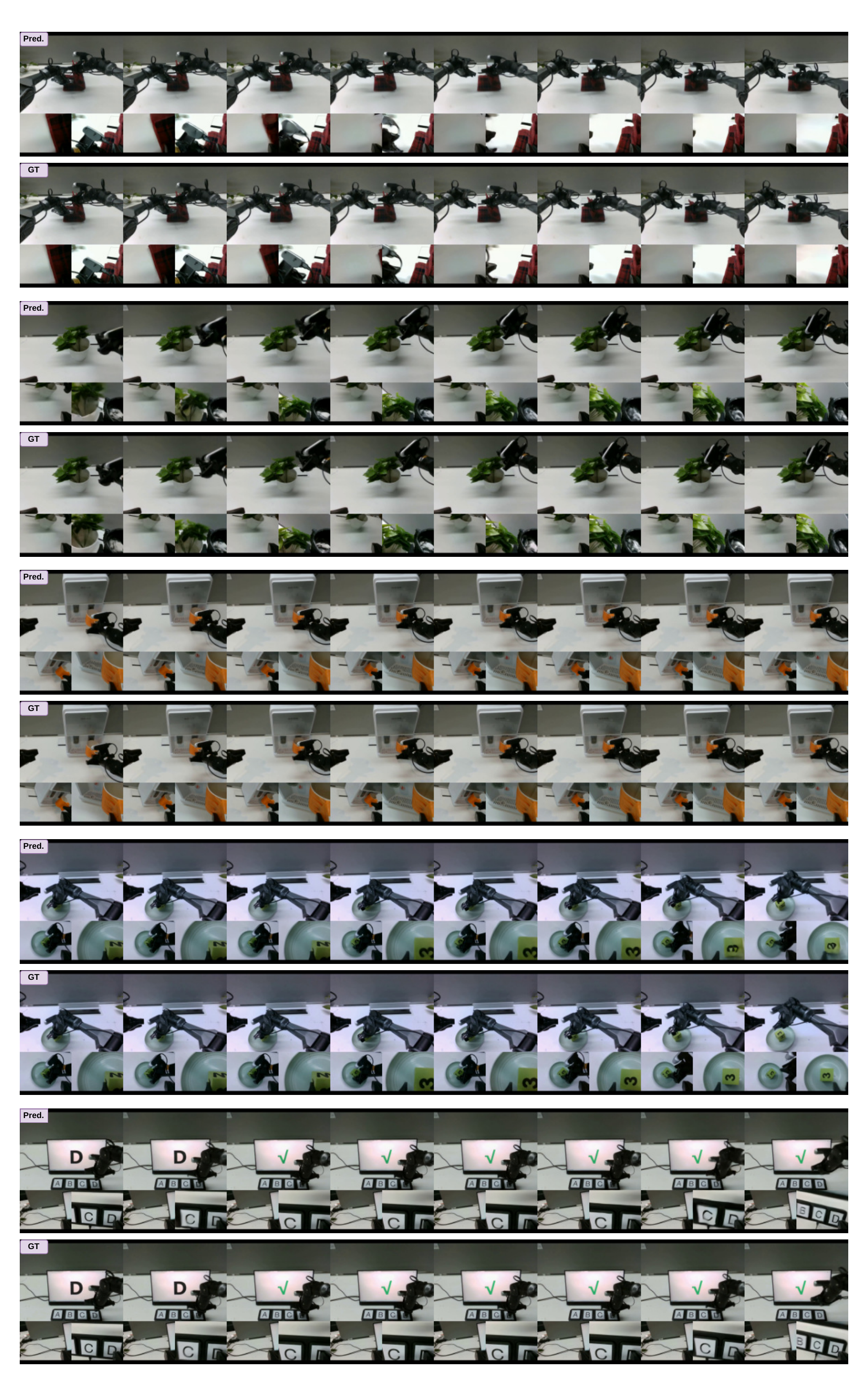}
    \caption{\textbf{Visualization of Motus's World Model Mode on Agilex-Aloha-2 Dataset.}}
    \label{fig:wm_aloha}
\vspace{-0.3cm}
\end{figure*}

\begin{figure*}[p!]
    \centering
    \includegraphics[width=0.8\linewidth]{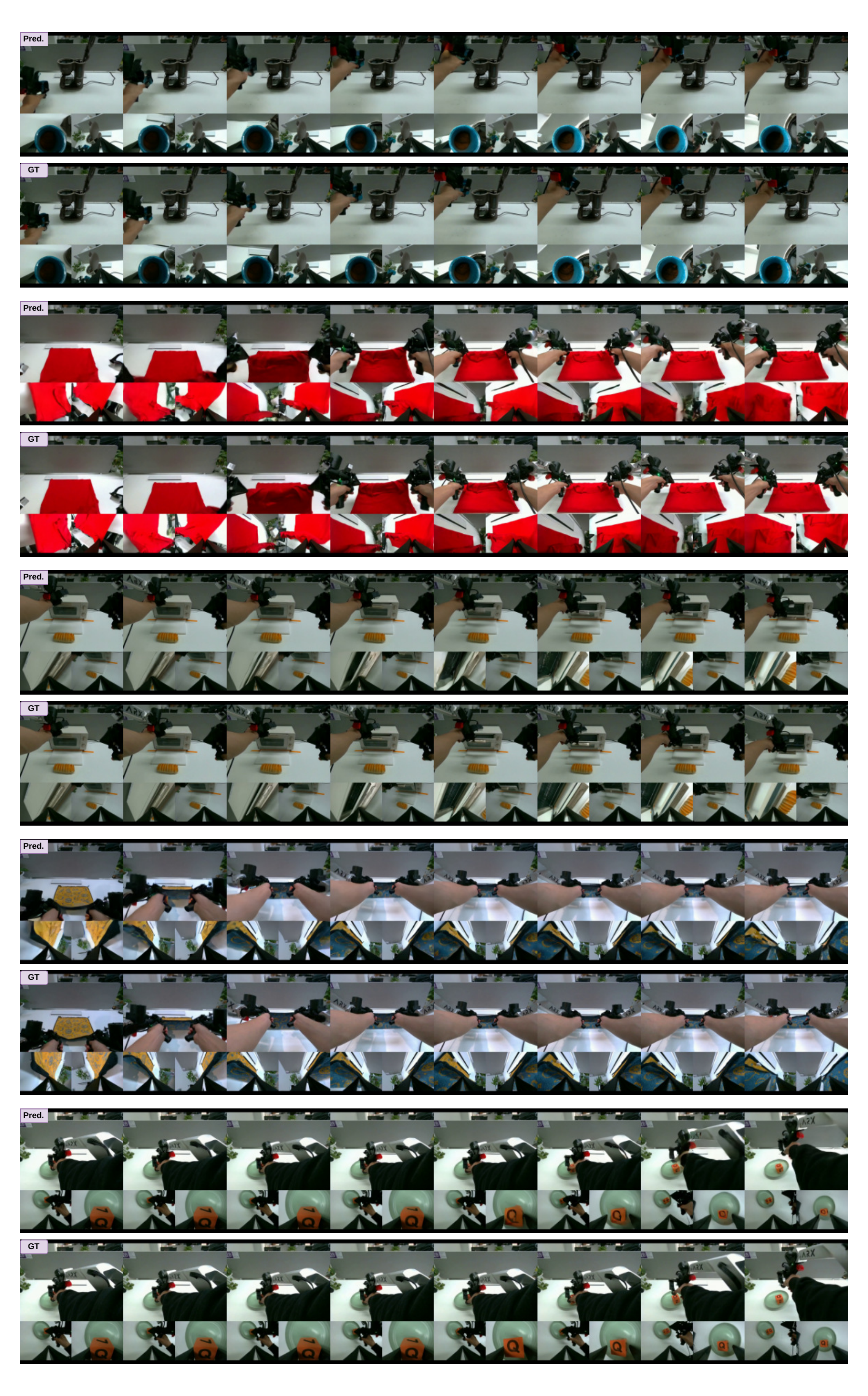}
    \caption{\textbf{Visualization of Motus's World Model Mode on AC-One Dataset.}}
    \label{fig:wm_acone}
\vspace{-0.3cm}
\end{figure*}

\paragraph{World Model.} 
As shown in Fig.~\ref{fig:wm_acone}, Fig.~\ref{fig:wm_aloha} and Tab.~\ref{tab:psr-questions}, when Motus performs in world model mode, it shows high-quality video generation results across two embodiments on real-world robot data, demonstrating strong future prediction capabilities.

\begin{table}[h]
\centering
  \footnotesize
\caption{\textbf{Generative Quality of Motus in World Model Mode}. The metrics were evaluated on real-world robot data across two robotic platform.}
\setlength{\tabcolsep}{4pt}
\begin{tabular}{cccccc}
\toprule
\textbf{Platform} & \textbf{FID}$\downarrow$ & \textbf{FVD}$\downarrow$ & \textbf{SSIM}$\uparrow$ & \textbf{LPIPS}$\downarrow$ & \textbf{PSNR}$\uparrow$ \\
\midrule
Agilex-Aloha-2 & 9.4571 & 49.2848 & 0.88618 & 0.05449 & 26.1021 \\
AC-One & 12.9609 & 73.1325 & 0.84605 & 0.07280 & 24.0379 \\
Avg. & 11.209 & 61.20865 & 0.8661 & 0.063645 & 25.0700 \\
\bottomrule
\end{tabular}
\label{tab:psr-questions}
\vspace{-0.3cm}
\end{table}

\paragraph{IDM.}

To validate the effectiveness of our model as an IDM, we trained two baseline IDMs for comparison: one based on a pretrained ResNet-18 backbone followed by an MLP layer, and another using DINOv2 features with an MLP head. Both models were trained on the RobotWin 2.0 randomized dataset using the Agilex-Aloha-2 robotic platform. Each model takes the current observation as input and predicts a sequence of future actions with an action chunk size of 16, which is consistent with the configuration used by Motus in RobotTwin. The training objective was to minimize the Mean Squared Error (MSE) between predicted and ground-truth actions.

As shown in Table~\ref{tab:idm}, when Motus performs in IDM mode, it achieves a lower action MSE than the specifically trained IDM baselines. This indicates that our model not only serves as an effective policy but also excels at inverse dynamics modeling, even outperforming models explicitly trained for that purpose.

\begin{table}[h]
\centering
  \footnotesize
\caption{\textbf{Action MSE of IDM}. The models are tested on 100 samples of RoboTwin 2.0 randomized data.}
\setlength{\tabcolsep}{4pt}
\begin{tabular}{ccc}
\toprule
\textbf{ResNet18+MLP} & \textbf{DINOv2+MLP} & \textbf{Motus} \\
\midrule
 0.044 & 0.122 & \textbf{0.014} \\
\bottomrule
\end{tabular}
\label{tab:idm}
\vspace{-0.3cm}
\end{table}

\noindent \textbf{VLA.} As shown in Tab.~\ref{tab:vla}, when Motus performs in the VLA mode, it also demonstrates competitive performance on RoboTwin 2.0 randomized data compared to the video-action joint prediction mode.
\vspace{0.5em}
\begin{table}[H]
\centering
  \footnotesize
\caption{\textbf{Average Success Rate on RoboTwin 2.0 Randomized Data of VLA.}}
\setlength{\tabcolsep}{4pt}

\begin{tabular}{ccc}
\toprule
\textbf{Motus (VLA)} & \textbf{Motus (Joint)} \\
\midrule
 83.90 & 87.02 \\
\bottomrule
\end{tabular}
\label{tab:vla}
\vspace{-0.3cm}
\end{table}

\begin{figure*}
    \centering
    \includegraphics[width=0.9\linewidth]{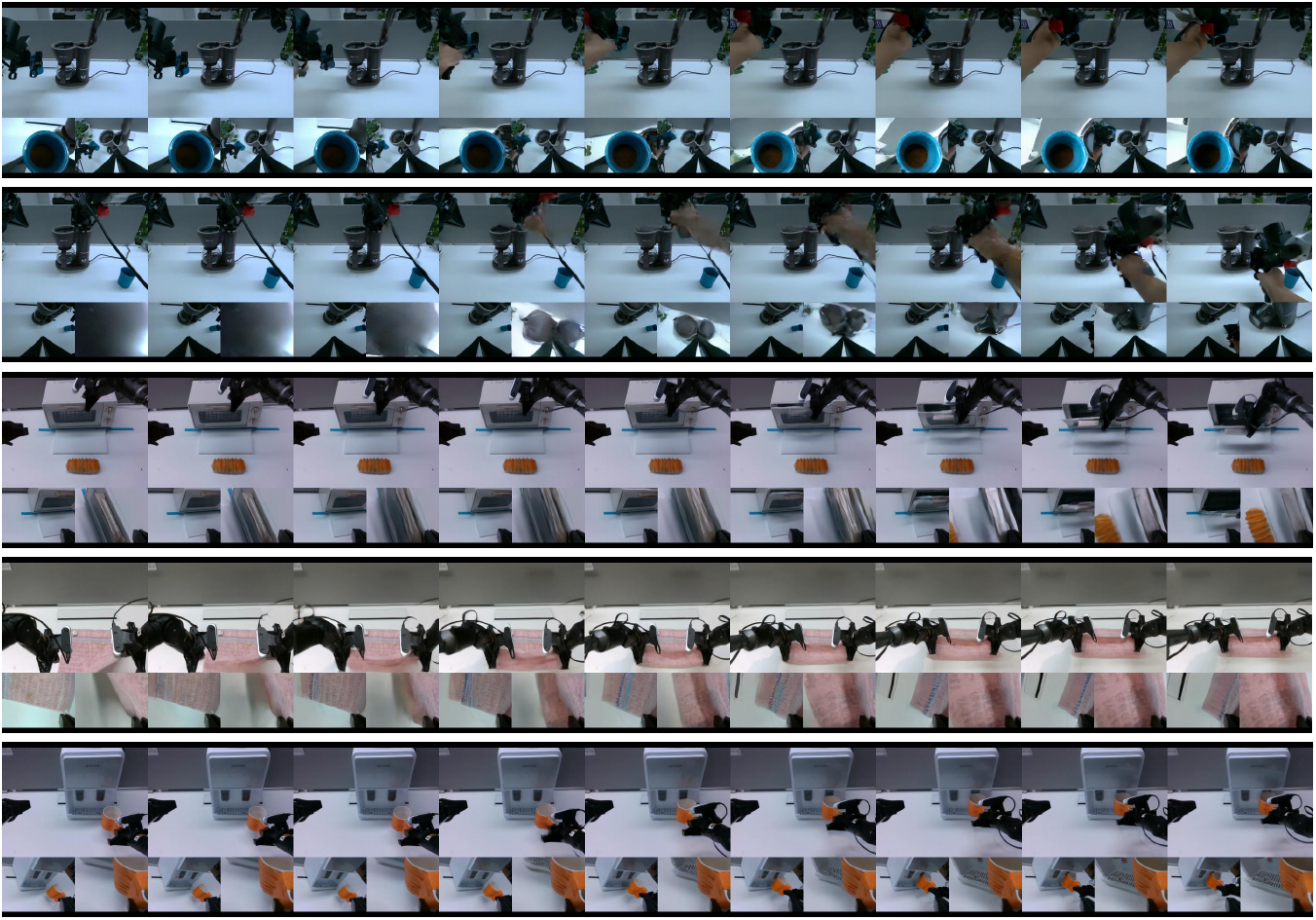}
    \caption{\textbf{Visualization of Motus’s Video-Action Joint Prediction Model mode during Real-World Inference.}}
    \label{fig:joint}
\vspace{-0.3cm}
\end{figure*}

\paragraph{Video-Action Joint Prediction Model.}
As shown in Fig.~\ref{fig:joint}, when Motus performs in the video-action joint prediction model mode, it demonstrates strong capabilities in generating both videos and precise actions simultaneously.

% \subsection{Video Generation}

% \subsection{World Models}

% \subsection{Inverse Dynamics}

% \subsection{Video-Action Joint Prediction}

% \subsection{Policy}

% \begin{figure*}
%     \centering
%     \includegraphics[width=0.9\linewidth]{figs/joint.drawio.pdf}
%     \caption{Caption}
%     \label{fig:placeholder}
% \vspace{-0.3cm}
% \end{figure*}

\section{More Experiments Results}

\subsection{Overall Comparison on RoboTwin 2.0 Simulation Data with More Baselines}

Tab.~\ref{tab:robotwin-full} shows the evaluation results on RoboTwin 2.0 Simulation, presenting the performance of Motus and other baselines on all 50 tasks under both clean scenes and randomized scenes.

\subsection{Other Benchmarks}
\paragraph{LIBERO-Long.} LIBERO-Long is the long-horizon subset of the LIBERO benchmark, comprising 10 language-conditioned manipulation tasks from LIBERO-100 that require multi-stage decision making, diverse manipulation skills, and robust knowledge transfer across objects and scenes. Under the standard LIBERO-Long evaluation protocol, our method achieves an average success score of \textbf{97.6}, matching the best reported performance of X-VLA and thereby reaching state-of-the-art results on this benchmark.

\begin{table}[h]
    \centering
    \footnotesize
    \begin{tabular}{cccccc}\toprule
         $\mathbf{\pi}_{\mathbf{0}}$ &  \textbf{GR00T-N1} & \textbf{UniVLA} & \textbf{OpenVLA-OFT} & \textbf{X-VLA} & \textbf{Motus} \\
    \midrule
        85.2 & 90.6 & 94.0 & 94.5 & 97.6 & \textbf{97.6} \\
    \bottomrule
    \end{tabular}
    \caption{Evaluation on LIBERO-Long Benchmark}
    \label{tab:placeholder}
\end{table}

\paragraph{VLABench.} VLAbench is an open-source benchmark for evaluating universal language-conditioned manipulation task learning, covering multiple dimensions such as manipulation skills, vision understanding, semantic comprehension, common sense, and reasoning. A single Motus model was fine-tuned on multiple tasks and subsequently evaluated based on its success rate across 3 tasks on 2 tracks provided by VLAbench: In Distribution and Cross Category. The result is shown in Tab.~\ref{tab:vlabench}. The evaluation result of $\pi_{0.5}$ is sourced from its official implementation.

\begin{table}[h]
    \centering
    \footnotesize
    \begin{tabular}{ccccc}\toprule
      Model & Add Condiment & Select Toy & Select Fruit & Avg. \\
    \midrule
    \multicolumn{5}{c}{\textbf{In Distribution}} \\
    \midrule
        $\pi_{0.5}$ & 0.56 & 0.3 & \textbf{0.42} & 0.43 \\
        \textbf{Motus} & \textbf{0.63} & \textbf{0.47} & 0.33 & \textbf{0.48}\\
    \midrule
    \multicolumn{5}{c}{\textbf{Cross Category}} \\
    \midrule
        $\pi_{0.5}$ & 0.06 & 0.24 & \textbf{0.36} & 0.22\\
        \textbf{Motus} & \textbf{0.14} & \textbf{0.40} & 0.20 & \textbf{0.25} \\
    \bottomrule
    \end{tabular}
    \caption{Evaluation of Success Rate on VLABench}
    \label{tab:vlabench}
\end{table}

\subsection{More Real-World Results}

\begin{figure}
  \centering
   \includegraphics[width=1\linewidth]{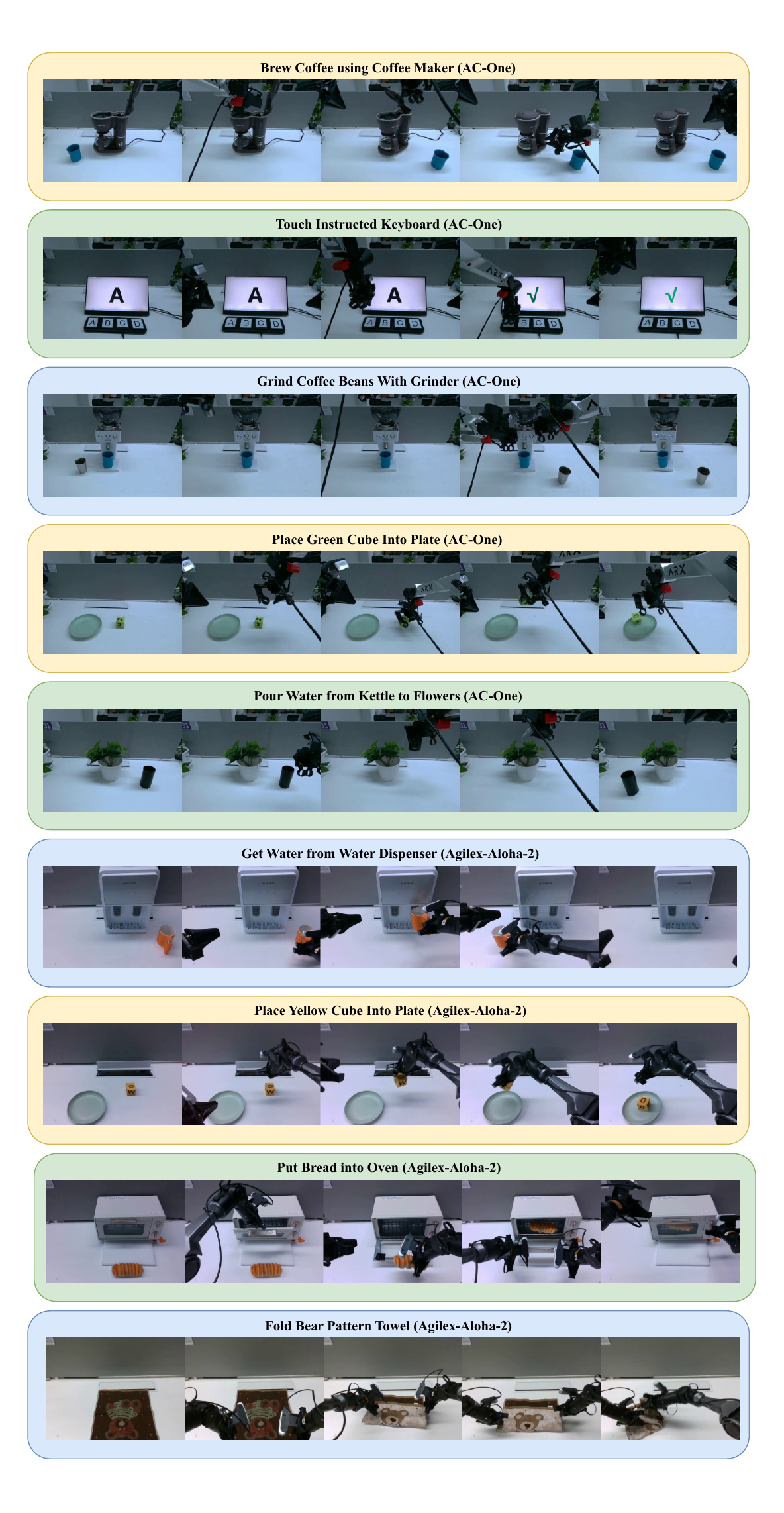}
   \caption{\textbf{Demonstrations of Motus for real-world tasks execution featuring 2 robots and 9 tasks.}}
   \label{fig:9tasks_demo}
\end{figure}

Fig.~\ref{fig:9tasks_demo} illustrates the visualization of the Motus execution for each task presented in Tab.~\ref{tab:real-world}. The detailed results containing subtask breakdown of the real-world tasks on the AC-One and Agilex-Aloha-2 platforms are presented in Tab.~\ref{tab:ac-one-full-result} and Tab.~\ref{tab:agilex-aloha-2-full-result}. The number preceding each subtask indicates the score assigned to its successful completion. For the towel-folding task, we evaluate each towel type four times. For the grab-cube task, we evaluate each cube type five times for both the in-domain and out-of-domain settings.

\noindent

\section{Implementation Details}
\subsection{Model Architecture}

Tab.~\ref{tab:architecture_details} provides the key hyperparameter settings for the Motus model architecture.

\begin{table}[t]
\centering
\small
\begin{tabular}{l|c}
\hline
\textbf{Component} & \textbf{Configuration} \\
\hline
\multicolumn{2}{l}{\textbf{Action Expert}} \\
Hidden Size & 1024 \\
Layers & 30 \\
Attention Heads & 24 \\
Layer Norm Epsilon & 1e-5 \\
Activation Function & GELU \\
\hline
\multicolumn{2}{l}{\textbf{Understand Expert}} \\
Hidden Size & 512 \\
Layers & 30 \\
Attention Heads & 24 \\
Layer Norm Epsilon & 1e-5 \\
Activation Function & GELU \\
\hline
\multicolumn{2}{l}{\textbf{Latent Action VAE}} \\
$\lambda_a$ (Action Alignment) & 1.0 \\
$\beta$ (KL Regularization) & $1\times10^{-6}$ \\
\hline
\multicolumn{2}{l}{\textbf{Sampling Rate}} \\
Video Frames & 8 @ 5Hz \\
Action Chunk & 48 @ 30Hz \\
\hline
\multicolumn{2}{l}{\textbf{Flow Matching}} \\
Inference Steps & 10 \\
Sampling Strategy & Logit Normal \\
\hline
\multicolumn{2}{l}{\textbf{Model Scale}} \\
VGM & ~5.00B \\
VLM & ~2.13B \\
Act. Expert & ~641.5M \\
Und. Expert & ~253.5M \\
Total & ~8B \\
\hline
\end{tabular}
\caption{Motus architecture hyperparameters and key configuration settings.}
\label{tab:architecture_details}
\end{table}

\subsection{Datasets}
\label{app:dataset_info}

Tab.~\ref{tab:app-datasets} shows the training data of Motus.

\begin{table*}[h]
  \caption{Detailed information about pre-training and fine-tuning datasets.}
  \label{tab:app-datasets}
  \centering
  \begin{tabular}{llll}
    \toprule
    Dataset    & Size & Embodiment & Data Level in the Pyramid \\
    \midrule
    Egodex~\cite{hoque2025egodex}    & 230,949   & Human & Level 2: Egocentric Human Videos\\
    Agibot~\cite{contributors2025agibotworld}    &  728,209  &  Genie-1 Robot &  Level 5: Multi-Robot Task Trajectory Data \\
    RDT~\cite{liu2024rdt}             &  6,083    &   Aloha Robot & Level 5: Multi-Robot Task Trajectory Data  \\
    RoboMind Franka~\cite{wu2024robomind} &   9,589  &  Franka Robot  &  Level 5: Multi-Robot Task Trajectory Data \\
    RoboMind Aloha~\cite{wu2024robomind} &  7,272    &  Aloha Robot  &  Level 5: Multi-Robot Task Trajectory Data \\
    % total: 982,102
    % \midrule
    RoboTwin~\cite{chen2025robotwin} & 27,500 & Aloha Robot & Level 3: Synthetic Data\\
    Task-Agnostic Data~\cite{tan2025anyposautomatedtaskagnosticactions} & 1,000 & Aloha Robot & Level 4: Task-Agnostic Data \\
    In-house Data            & 2,000       &  Aloha Robot &  Level 6: Target-Robot Task Trajectory Data \\
    % total: 984,409
    \bottomrule
  \end{tabular}
\end{table*}

\subsection{Training Configuration}
 Tab.~\ref{tab:training_config_three_stages} provides the detailed training configuration for the three stages of Motus.
\begin{table*}[t]
\centering
\footnotesize
\setlength{\tabcolsep}{6pt}
\renewcommand{\arraystretch}{1.2}
\caption{\textbf{Training Configuration across Three Stages}.}
\begin{tabular}{l|c|c|c}
\toprule
\textbf{Stages} & \textbf{Stage 1} & \textbf{Stage 2} & \textbf{Stage 3} \\
\textbf{Batch Size} & 256 & 256 & 256 \\
\textbf{Learning Rate} & $8 \times 10^{-5}$ & $5 \times 10^{-5}$ & $1\sim5 \times 10^{-5}$ \\
\textbf{Optimizer} & AdamW & AdamW & AdamW \\
\textbf{Weight Decay} & 0.01 & 0.01 & 0.01 \\
\textbf{GPU Hours} & \textasciitilde8000  & \textasciitilde10000 & \textasciitilde400  \\
\bottomrule
\end{tabular}
\label{tab:training_config_three_stages}
\end{table*}

\begin{table*}[ht]
  \centering
  \footnotesize
  \setlength{\tabcolsep}{5.5pt}
  \caption{\textbf{Evaluation on RoboTwin 2.0 Simulation (Clean vs Randomized, 50+ tasks).}}
  \begin{tabular}{*{1}{>{\centering\arraybackslash}m{3.0cm}} *{12}{>{\centering\arraybackslash}m{0.80cm}}}
    \toprule
    \textbf{\makecell[c]{Simulation Task}} 
      & \multicolumn{2}{c}{\textbf{GO-1}}
      & \multicolumn{2}{c}{$\mathbf{\pi}_{\mathbf{0.5}}$}
      & \multicolumn{2}{c}{\textbf{X-VLA}}
      & \multicolumn{2}{c}{\textbf{w/o Pretrain}}
      & \multicolumn{2}{c}{\textbf{Stage1}}
      & \multicolumn{2}{c}{\textbf{Motus}} \\
    & \textbf{Clean} & \textbf{Rand.} 
    & \textbf{Clean} & \textbf{Rand.} 
    & \textbf{Clean} & \textbf{Rand.} 
    & \textbf{Clean} & \textbf{Rand.} 
    & \textbf{Clean} & \textbf{Rand.} 
    & \textbf{Clean} & \textbf{Rand.} \\
    \midrule
\textit{Adjust Bottle} & 49\% & 62\% & 79\% & 83\% & \textbf{100\%} & \textbf{99\%} & 99\% & 97\% & 98\% & 94\% & 89\% & 93\% \\
\textit{Beat Block Hammer} & 6\% & 10\% & 63\% & 50\% & 92\% & 88\% & 88\% & \textbf{90\%} & 88\% & 82\% & \textbf{95\%} & 88\% \\
\textit{Blocks Ranking Rgb} & 7\% & 3\% & 43\% & 35\% & 83\% & 83\% & 92\% & 88\% & 97\% & \textbf{98\%} & \textbf{99\%} & 97\% \\
\textit{Blocks Ranking Size} & 2\% & 2\% & 8\% & 14\% & 67\% & \textbf{74\%} & 38\% & 50\% & 73\% & 68\% & \textbf{75\%} & 63\% \\
\textit{Click Alarmclock} & 95\% & 90\% & 97\% & 93\% & 99\% & 99\% & 100\% & 99\% & \textbf{100\%} & \textbf{100\%} & \textbf{100\%} & \textbf{100\%} \\
\textit{Click Bell} & 98\% & 95\% & 75\% & 76\% & 100\% & 100\% & 100\% & 100\% & \textbf{100\%} & \textbf{100\%} & \textbf{100\%} & \textbf{100\%} \\
\textit{Dump Bin Bigbin} & 57\% & 45\% & 30\% & 42\% & 79\% & 77\% & 94\% & \textbf{96\%} & \textbf{98\%} & \textbf{96\%} & 95\% & 91\% \\
\textit{Grab Roller} & 99\% & 99\% & 90\% & 89\% & 100\% & 100\% & 100\% & 100\% & \textbf{100\%} & \textbf{100\%} & \textbf{100\%} & \textbf{100\%} \\
\textit{Handover Block} & 9\% & 12\% & 18\% & 19\% & 73\% & 37\% & 34\% & 15\% & 55\% & 55\% & \textbf{86\%} & \textbf{73\%} \\
\textit{Handover Mic} & 12\% & 8\% & 28\% & 18\% & 0\% & 0\% & \textbf{98\%} & \textbf{95\%} & 80\% & 88\% & 78\% & 63\% \\
\textit{Hanging Mug} & 0\% & 0\% & 3\% & 3\% & 23\% & 27\% & 14\% & 10\% & 37\% & 25\% & \textbf{38\%} & \textbf{38\%} \\
\textit{Lift Pot} & 92\% & 92\% & 0\% & 0\% & \textbf{99\%} & \textbf{100\%} & 90\% & 87\% & 87\% & 84\% & 96\% & 99\% \\
\textit{Move Can Pot} & 16\% & 4\% & 29\% & 27\% & \textbf{89\%} & \textbf{86\%} & 43\% & 53\% & 56\% & 65\% & 34\% & 74\% \\
\textit{Move Pillbottle Pad} & 9\% & 11\% & 33\% & 29\% & 73\% & 71\% & 83\% & 83\% & \textbf{96\%} & 90\% & 93\% & \textbf{96\%} \\
\textit{Move Playingcard Away} & 37\% & 24\% & 59\% & 67\% & 93\% & \textbf{98\%} & 50\% & 47\% & 77\% & 84\% & \textbf{100\%} & 96\% \\
\textit{Move Stapler Pad} & 3\% & 4\% & 16\% & 18\% & 78\% & 73\% & 49\% & 37\% & 75\% & 68\% & \textbf{83\%} & \textbf{85\%} \\
\textit{Open Laptop} & 65\% & 60\% & 19\% & 35\% & 93\% & \textbf{100\%} & 89\% & 89\% & 91\% & 96\% & \textbf{95\%} & 91\% \\
\textit{Open Microwave} & 12\% & 14\% & 35\% & 37\% & 79\% & 71\% & 83\% & 82\% & 82\% & 84\% & \textbf{95\%} & \textbf{91\%} \\
\textit{Pick Diverse Bottles} & 61\% & 56\% & 5\% & 3\% & 58\% & 36\% & 53\% & 62\% & 18\% & 18\% & \textbf{90\%} & \textbf{91\%} \\
\textit{Pick Dual Bottles} & 81\% & 74\% & 10\% & 6\% & 47\% & 36\% & 58\% & 68\% & 7\% & 17\% & \textbf{96\%} & \textbf{90\%} \\
\textit{Place A2b Left} & 33\% & 36\% & 62\% & 60\% & 48\% & 49\% & 78\% & 79\% & \textbf{93\%} & \textbf{82\%} & 88\% & 79\% \\
\textit{Place A2b Right} & 31\% & 22\% & 62\% & 57\% & 36\% & 36\% & 86\% & 83\% & \textbf{94\%} & \textbf{90\%} & 91\% & 87\% \\
\textit{Place Bread Basket} & 47\% & 52\% & 48\% & 56\% & 81\% & 71\% & 73\% & 83\% & 89\% & 87\% & \textbf{91\%} & \textbf{94\%} \\
\textit{Place Bread Skillet} & 2\% & 1\% & 38\% & 46\% & 77\% & 67\% & 71\% & 71\% & \textbf{86\%} & \textbf{87\%} & \textbf{86\%} & 83\% \\
\textit{Place Burger Fries} & 88\% & 92\% & 66\% & 70\% & 94\% & 94\% & 95\% & 90\% & 97\% & \textbf{99\%} & \textbf{98\%} & 98\% \\
\textit{Place Can Basket} & 29\% & 37\% & 19\% & 25\% & 49\% & 52\% & 46\% & 62\% & 66\% & 55\% & \textbf{81\%} & \textbf{76\%} \\
\textit{Place Cans Plasticbox} & 68\% & 77\% & 40\% & 47\% & 97\% & 98\% & 96\% & 99\% & 97\% & \textbf{100\%} & \textbf{98\%} & 94\% \\
\textit{Place Container Plate} & 73\% & 70\% & 71\% & 78\% & 97\% & 95\% & 97\% & \textbf{100\%} & \textbf{98\%} & 98\% & \textbf{98\%} & 99\% \\
\textit{Place Dual Shoes} & 6\% & 10\% & 12\% & 7\% & 79\% & 88\% & 78\% & 80\% & \textbf{94\%} & \textbf{94\%} & 93\% & 87\% \\
\textit{Place Empty Cup} & 44\% & 39\% & 75\% & 86\% & \textbf{100\%} & \textbf{98\%} & 97\% & 97\% & 96\% & 97\% & 99\% & \textbf{98\%} \\
\textit{Place Fan} & 1\% & 0\% & 25\% & 36\% & 80\% & 75\% & 77\% & 85\% & 77\% & 85\% & \textbf{91\%} & \textbf{87\%} \\
\textit{Place Mouse Pad} & 15\% & 10\% & 21\% & 26\% & 70\% & \textbf{70\%} & 62\% & 68\% & \textbf{72\%} & 69\% & 66\% & 68\% \\
\textit{Place Object Basket} & 48\% & 49\% & 43\% & 36\% & 44\% & 39\% & 74\% & 75\% & 76\% & 80\% & \textbf{81\%} & \textbf{87\%} \\
\textit{Place Object Scale} & 26\% & 27\% & 40\% & 49\% & 52\% & 74\% & 84\% & 83\% & \textbf{88\%} & \textbf{93\%} & \textbf{88\%} & 85\% \\
\textit{Place Object Stand} & 56\% & 63\% & 74\% & 65\% & 86\% & 88\% & 91\% & 93\% & 93\% & 96\% & \textbf{98\%} & \textbf{97\%} \\
\textit{Place Phone Stand} & 30\% & 37\% & 49\% & 53\% & \textbf{88\%} & \textbf{87\%} & 80\% & 78\% & 76\% & 86\% & 87\% & 86\% \\
\textit{Place Shoe} & 15\% & 13\% & 57\% & 61\% & 96\% & 95\% & 95\% & 92\% & \textbf{100\%} & \textbf{99\%} & 99\% & 97\% \\
\textit{Press Stapler} & 66\% & 51\% & 80\% & 70\% & 92\% & 98\% & \textbf{97\%} & 94\% & 96\% & \textbf{98\%} & 93\% & \textbf{98\%} \\
\textit{Put Bottles Dustbin} & 7\% & 4\% & 12\% & 9\% & 74\% & 77\% & 36\% & 33\% & 34\% & 24\% & \textbf{81\%} & \textbf{79\%} \\
\textit{Put Object Cabinet} & 60\% & 43\% & 24\% & 15\% & 46\% & 48\% & 84\% & 64\% & \textbf{97\%} & \textbf{87\%} & 88\% & 71\% \\
\textit{Rotate Qrcode} & 22\% & 9\% & 47\% & 56\% & 34\% & 33\% & 80\% & 60\% & \textbf{91\%} & \textbf{79\%} & 89\% & 73\% \\
\textit{Scan Object} & 1\% & 2\% & 42\% & 38\% & 14\% & 36\% & 42\% & 50\% & 56\% & \textbf{69\%} & \textbf{67\%} & 66\% \\
\textit{Shake Bottle Horizontally} & 97\% & 92\% & 96\% & \textbf{100\%} & 100\% & \textbf{100\%} & 100\% & 97\% & \textbf{100\%} & 96\% & \textbf{100\%} & 98\% \\
\textit{Shake Bottle} & 97\% & 93\% & 91\% & \textbf{100\%} & 99\% & \textbf{100\%} & \textbf{100\%} & 96\% & 99\% & 97\% & \textbf{100\%} & 97\% \\
\textit{Stack Blocks Three} & 1\% & 1\% & 15\% & 16\% & 6\% & 10\% & 71\% & 76\% & \textbf{99\%} & \textbf{95\%} & 91\% & \textbf{95\%} \\
\textit{Stack Blocks Two} & 12\% & 22\% & 48\% & 56\% & 92\% & 87\% & 96\% & 94\% & 99\% & \textbf{99\%} & \textbf{100\%} & 98\% \\
\textit{Stack Bowls Three} & 4\% & 7\% & 33\% & 35\% & 76\% & 86\% & \textbf{90\%} & 74\% & 86\% & 83\% & 79\% & \textbf{87\%} \\
\textit{Stack Bowls Two} & 51\% & 45\% & 78\% & 66\% & 96\% & 93\% & \textbf{98\%} & 98\% & 97\% & \textbf{98\%} & \textbf{98\%} & \textbf{98\%} \\
\textit{Stamp Seal} & 19\% & 13\% & 36\% & 23\% & 76\% & 82\% & 80\% & 88\% & \textbf{93\%} & \textbf{95\%} & \textbf{93\%} & 92\% \\
\textit{Turn Switch} & 34\% & 30\% & 5\% & 6\% & 40\% & 61\% & 69\% & 60\% & 59\% & 64\% & \textbf{84\%} & \textbf{78\%} \\
\midrule
\textbf{\textit{Average (\%)}} & 37.8 & 36.24 & 42.98 & 43.84 & 72.8 & 72.84 & 77.56 & 77.00 & 82.26 & 81.86 & \textbf{88.66} & \textbf{87.02} \\
    \bottomrule
  \end{tabular}
  \label{tab:robotwin-full}
\vspace{-0.3cm}
\end{table*}

\begin{table}[t]
\centering
\footnotesize
\caption{Real-World Tasks on AC-One Platform with a Detailed Subtask Breakdown.}
\setlength{\tabcolsep}{4pt}

\begin{tabular}{lccc}
\toprule
\textbf{Subgoal} & $\pi_{0.5}$ & \textbf{w/o Pretrain} & \textbf{Motus} \\
\midrule
\multicolumn{4}{c}{\textbf{Fold Towel}} \\
\multicolumn{4}{c}{\textbf{Types: bear-pattern/blue-yellow/purple/red-blue/pink}} \\
\midrule
0.0: Complete Failure      & 16 & 19 & 13 \\
0.2: Grab both sides       & 4  & 1  & 3  \\
0.5: One fold complete     & -  & -  & 3  \\
0.8: Grab the right side   & -  & - & 1  \\
1.0: Two folds complete    & -  & -  & -  \\
Partial Success Rate     & 4\%  & 1\%  & 14.5\% \\
\midrule
\multicolumn{4}{c}{\textbf{Grab Cube}} \\
\multicolumn{4}{c}{\textbf{Types: red/orange/green/yellow}} \\
\midrule
0.0: Complete Failure      & 7 & 8 & - \\
0.5: Grab cube             & 3  & -  & - \\
1.0: Put cube into plate   & 10  & 12  & 20 \\
Partial Success Rate    & 57.5\%  & 60\%  & 100\% \\
\midrule
\multicolumn{4}{c}{\textbf{Grab Cube}} \\
\multicolumn{4}{c}{\textbf{OOD setting: cube placed outside training space}} \\
\midrule
0.0: Complete Failure      & 11 & 13 & 4 \\
0.5: Grab cube             & 1  & -  & - \\
1.0: Put cube into plate   & 4  & 3  & 12 \\
Partial Success Rate    & 28.125\%  & 18.75\%  & 75\% \\
\midrule
\multicolumn{4}{c}{\textbf{Brew Coffee using Drip Coffee Machine}} \\
\midrule
0.0: Complete Failure      & 10 & 10 & 2 \\
0.2: Grab the blue cup     & - & -  & 1 \\
0.5: Pour coffee grounds     & -  & -  & - \\
0.8:  Close the lid     & -  & -  & 5 \\
1.0: Turn on the switch   & -  & -  & 2 \\
Partial Success Rate    & 0\%  & 0\%  & 62\% \\
\midrule
\multicolumn{4}{c}{\textbf{Get Water from Water Dispenser}} \\
\midrule
0.0: Complete Failure      & 4 & 9 & 4 \\
0.4: Grab the orange cup     & 5  & - & 4 \\
0.8: Fill the cup with water     & -  & 1  & - \\
1.0: Put down the cup   & 1 & -  & 2 \\
Partial Success Rate    & 30\%  & 8\%  & 36\% \\
\midrule
\multicolumn{4}{c}{\textbf{Grind Coffee Beans with Grinder}} \\
\midrule
0.0: Complete Failure      & 9 & 10 & - \\
0.3: Grab the metal cup     & -  & -  & - \\
0.8: Pour the coffee beans     & 1  & -  & 4 \\
1.0: Press the button   & -  & -  & 6 \\
Partial Success Rate    & 8\%  & 0\% & 92\% \\
\midrule
\multicolumn{4}{c}{\textbf{Pour Water from Kettle to Flowers}} \\
\midrule
0.0: Complete Failure      & 18 & 18 & 4 \\
0.5: Grab the black cup     & 2  & 2 & 6 \\
1.0: Pour water   & -  & -  & 10 \\
Partial Success Rate    & 5\%  & 5\%  & 65\% \\
\midrule
\multicolumn{4}{c}{\textbf{Touch Keyboard with Hand for Multiple Choice Questions}} \\
\midrule
0.0: Complete Failure      & 20 & - & 3 \\
0.5: Use the correct arm     & - & - & 1 \\
1.0: Press the right key   & -  & 20  & 16 \\
Partial Success Rate    & 0\%  & 100\%  & 82.5\% \\
\bottomrule
\label{tab:ac-one-full-result}
\end{tabular}
\end{table}

\begin{table}[t]
\centering
\footnotesize
\caption{Real-World Tasks on Agilex-Aloha-2 Platform with a Detailed Subtask Breakdown.}
\setlength{\tabcolsep}{4pt}

\begin{tabular}{lccc}
\toprule
\textbf{Subgoal} & $\pi_{0.5}$ & \textbf{w/o Pretrain} & \textbf{Motus} \\
\midrule
\multicolumn{4}{c}{\textbf{Fold Towel}} \\
\multicolumn{4}{c}{\textbf{Types: bear-pattern/blue-yellow/purple/red-blue/pink}} \\
\midrule
0.0: Complete Failure      & 4 & 20 & 5 \\
0.2: Grab both sides       & 11  & -  & 1  \\
0.5: One fold complete     & 3  & -  & 12  \\
0.8: Grab the right side   & 1  & - & 2  \\
1.0: Two folds complete    & 1  & -  & -  \\
Partial Success Rate    & 27.5\%  & 0\%  & 39\% \\
\midrule
\multicolumn{4}{c}{\textbf{Grab Cube}} \\
\multicolumn{4}{c}{\textbf{Types: red/orange/green/yellow}} \\
\midrule
0.0: Complete Failure      & 2 & 8 & - \\
0.5: Grab cube             & 1  & 8  & - \\
1.0: Put cube into plate   & 17  & 4  & 20 \\
Partial Success Rate    & 87.5\%  & 40\%  & 100\% \\
\midrule
\multicolumn{4}{c}{\textbf{Grab Cube}} \\
\multicolumn{4}{c}{\textbf{OOD setting: cube placed outside training space}} \\
\midrule
0.0: Complete Failure      & 5 & 13 & 11 \\
0.5: Grab cube             & -  & -  & - \\
1.0: Put cube into plate   & 11  & 3  & 5 \\
Partial Success Rate    & 68.75\%  & 18.75\%  & 31.25\% \\
\midrule
\multicolumn{4}{c}{\textbf{Put Bread into Oven }} \\
\midrule
0.0: Complete Failure      & 5 & 10 & 5 \\
0.2: Open the oven     & -  & -  & - \\
0.4: Grab the bread     & 1  & -  & - \\
0.6: Put the bread into the oven   & - & -  & 3 \\
0.8: Close the oven   & 4 & -  & 2 \\
1.0: Spin the button   & - & -  & - \\
Partial Success Rate    & 36\%  & 0\%  & 34\% \\
\midrule
\multicolumn{4}{c}{\textbf{Pour Water from Kettle to Flowers}} \\
\midrule
0.0: Complete Failure      & 2 & 4 & 3 \\
0.5: Grab the black cup     & 18  & 16 & 15 \\
1.0: Pour water   & -  & -  & 2 \\
Partial Success Rate    & 45\%  & 40\%  & 47.5\% \\
\midrule
\multicolumn{4}{c}{\textbf{Touch Keyboard with Hand for Multiple Choice Questions}} \\
\midrule
0.0: Complete Failure      & 5 & - & - \\
0.5: Use the correct arm     & 1 & 6 & 8 \\
1.0: Press the right key   & 14  & 14  & 12 \\
Partial Success Rate     & 72.5\%  & 85\%  & 80\% \\
\bottomrule
\label{tab:agilex-aloha-2-full-result}
\end{tabular}
\end{table}

\clearpage   

\end{document}